\documentclass{article}

\usepackage{microtype}
\usepackage{graphicx}
\usepackage{float} %
\usepackage{subcaption}
\usepackage{booktabs} %

\usepackage{hyperref}

\usepackage[accepted]{icml2026}

\usepackage{amsmath}
\usepackage{amssymb}
\usepackage{mathtools}
\usepackage{amsthm}
\usepackage{tabularx,makecell}
\usepackage{dsfont} %
\usepackage{xcolor} %
\usepackage{pifont} %

\usepackage{hyperref}
\usepackage{url}

\def\scheme{\textsc{FLIPS}\xspace}
\def\extraction{\text{extraction}\xspace}

\def\reference{\text{reference}\xspace}

\usepackage{algorithm}
\usepackage{algorithmic}
\newcommand{\RETURN}{\STATE \textbf{Return} }

\usepackage{color} %
\usepackage{tcolorbox}
\usepackage{xspace}
\usepackage{listings}
\usepackage{enumitem}
\usepackage{booktabs}
\usepackage{multirow}
\usepackage{threeparttable}
\usepackage{makecell}

\usepackage{subcaption}
\usepackage[capitalize,noabbrev]{cleveref}

\theoremstyle{plain}

\theoremstyle{definition}

\theoremstyle{remark}

\usepackage[textsize=tiny]{todonotes}

\usepackage{amsmath,amsfonts,bm}

\def\eqref#1{equation~\ref{#1}}

\def\1{\bm{1}}

\DeclareMathAlphabet{\mathsfit}{\encodingdefault}{\sfdefault}{m}{sl}
\SetMathAlphabet{\mathsfit}{bold}{\encodingdefault}{\sfdefault}{bx}{n}

\DeclareMathOperator*{\argmax}{arg\,max}

\definecolor{ForestGreen}{RGB}{34,139,34}
\definecolor{BrickRed}{RGB}{178,34,34}
\usepackage{graphicx}

\newcommand{\nNPairs}{31}

\newcommand{\nClosedEight}{60.1\%}

\newcommand{\nClosedThirtyTwo}{76.6\%}
\newcommand{\nClosedSixtyFour}{82.6\%}

\newcommand{\nLLMmapFour}{32.3\%}
\newcommand{\nLLMmapBest}{33.4\%}
\newcommand{\nLLMmapBestNt}{6}

\newcommand{\nAucThirtyTwo}{0.684}

\newcommand{\nCfiftyAccThirtyTwo}{65.9\%}

\newcommand{\nRecCfiftyThirtyTwo}{41.1\%}
\newcommand{\nPrecCfiftyThirtyTwo}{22.9\%}

\newcommand{\nAttrRelModel}{1\%}
\newcommand{\nAttrRelQuant}{19\%}
\newcommand{\nAttrRelT}{5\%}
\newcommand{\nAttrRelSP}{25\%}
\newcommand{\nAttrMissSP}{16.8\%}
\newcommand{\nAttrRandSP}{67.4\%}
\newcommand{\nUnseenPrev}{9.6\%}

\newcommand{\nAbsClosedThirtyTwo}{77\%}

\newcommand{\nAbsLLMmapBest}{33\%}
\newcommand{\nAbsRecCfifty}{41\%}
\newcommand{\nAbsPrecCfifty}{23\%}
\newcommand{\nPrecLiftCfifty}{2.4}

\newcommand{\nFourNistOthers}{0.46}
\newcommand{\nFourNistZeroOne}{0.74}
\newcommand{\nFourRhoAll}{-0.06}

\newcommand{\nFourRecallOthers}{76.1\%}
\newcommand{\nFourRecallZeroOne}{75.5\%}

\newcommand{\nFourShareBeatsAll}{81\%}
\newcommand{\nHeatGrand}{76.2\%}
\newcommand{\nHeatLlmLow}{59.0\%}
\newcommand{\nHeatLlmLowName}{Qwen2-1.5B-Instruct}
\newcommand{\nHeatLlmHigh}{91.6\%}
\newcommand{\nHeatLlmHighName}{gemma-2-27b-it}
\newcommand{\nHeatVarLow}{55.0\%}
\newcommand{\nHeatVarHigh}{97.7\%}
\newcommand{\nHeatSpTwo}{86.2\%}
\newcommand{\nHeatSpThree}{57.3\%}
\newcommand{\nHeatSpFour}{55.0\%}
\newcommand{\nHeatTempZeroFour}{97.7\%}
\newcommand{\nHeatTempOneZero}{71.5\%}
\newcommand{\nHeatFpEight}{60.7\%}
\newcommand{\nHeatIntFour}{86.6\%}
\newcommand{\nHeatAbli}{75.4\%}
\newcommand{\nHeatQwenSeventyTwoSpTwo}{100\%}
\newcommand{\nHeatQwenThirtySpTwo}{100\%}
\newcommand{\nImpTop}{4.4\%}
\newcommand{\nImpShareNist}{54\%}
\newcommand{\nImpSharePast}{28\%}
\newcommand{\nImpShareOpen}{14\%}
\newcommand{\nImpShareLen}{4\%}

\newcommand{\nTruncFullThirtyTwo}{75.7\%}
\newcommand{\nTruncHundThirtyTwo}{71.4\%}

\newcommand{\nPRRecallPNineNine}{42\%}

\newcommand{\nAbsDistOne}{90.9\%}

\newcommand{\rPubClosedEight}{0.9621} %
\newcommand{\rNanRowsLo}{0.556} %
\newcommand{\rNanRowsHi}{0.937} %
\newcommand{\rNanRowsMean}{0.766} %
\newcommand{\rCellsFilledLo}{0.02} %
\newcommand{\rCellsFilledHi}{0.03} %
\newcommand{\rMedianMissing}{1} %
\newcommand{\rZeroOneAcc}{0.33} %
\newcommand{\rThirtyPairMean}{0.681} %
\newcommand{\rDiscardAll}{0.0223} %
\newcommand{\rDiscardTopA}{0.1135} %
\newcommand{\rDiscardTopB}{0.1045} %
\newcommand{\rDiscardTopC}{0.0578} %
\newcommand{\rDiscardTone}{0.96} %
\expandafter\newcommand\csname rKlit1\endcsname{210}\expandafter\newcommand\csname rKlin1\endcsname{232}
\expandafter\newcommand\csname rKlit2\endcsname{187}\expandafter\newcommand\csname rKlin2\endcsname{227}
\expandafter\newcommand\csname rKlit3\endcsname{167}\expandafter\newcommand\csname rKlin3\endcsname{222}
\expandafter\newcommand\csname rKlit5\endcsname{131}\expandafter\newcommand\csname rKlin5\endcsname{212}
\expandafter\newcommand\csname rKlit8\endcsname{94}\expandafter\newcommand\csname rKlin8\endcsname{199}
\expandafter\newcommand\csname rKlit12\endcsname{60}\expandafter\newcommand\csname rKlin12\endcsname{182}
\expandafter\newcommand\csname rKlit23\endcsname{22}\expandafter\newcommand\csname rKlin23\endcsname{142}

\usepackage{xfp}
\usepackage{siunitx}
\newcommand{\fx}[2]{\num[round-precision=#1]{\fpeval{#2}}}
\newcommand{\pct}[1]{\fx{1}{100*(#1)}\%}
\newcommand{\pcti}[1]{\fx{0}{100*(#1)}\%}

\newcommand{\dec}[1]{\fx{3}{#1}}

\newcommand{\fgNs}{300}\newcommand{\fgNt}{32}\newcommand{\vNPairs}{31}
\newcommand{\fgEROnePaperNoLeakEight}{0.24568103111852746}
\newcommand{\ppsgn}[1]{\ifnum\fpeval{(#1)>=0}=1 +\fi\fx{1}{#1}\,pp}
\newcommand{\ppse}[2]{\ppsgn{#1} $\pm$ \fx{1}{#2}}
\newcommand{\xColsDelta}{4.304}
\newcommand{\xColsDeltaSe}{0.181}
\newcommand{\xColsDeltaEight}{4.932}

\newcommand{\fgAttrClasses}{237}

\newcommand{\fgAttrNt}{32}

\newcommand{\capAttrZ}{How often does \scheme miss each attribute, relative to random? For each attribute of an instance, we count the closed-set samples for which the instance predicted by \scheme has a different value of this attribute than the true instance, and divide by the same count for an instance drawn at random among the \fgAttrClasses{}: $0\%$ means that the attribute is never missed, $100\%$ that it is missed as often as at random. For example, \scheme misses the system prompt for \nAttrMissSP{} of the samples, against \nAttrRandSP{} at random, hence \nAttrRelSP{}. Closed-set setting, instance classifier of Section~\ref{sec:flips_scheme}, $N_t{=}\fgAttrNt$.}

\newcommand{\nDtwAucUsualThirtyTwo}{0.6875}

\newcommand{\nDtwAucDtwoThirtyTwo}{0.8505}

\def\scheme{\textsc{FLIPS}\xspace}

\icmltitlerunning{FLIPS: Instance-Fingerprinting for LLMs via Pseudo-random Sequences}
\begin{document}

\twocolumn[
  \icmltitle{FLIPS: Instance-Fingerprinting for LLMs via Pseudo-random Sequences}

  \icmlsetsymbol{equal}{*}

\begin{icmlauthorlist}
    \icmlauthor{Gurvan Richardeau}{peren}
    \icmlauthor{Gohar Dashyan}{peren}
    \icmlauthor{Erwan {Le Merrer}}{inria}
    \icmlauthor{Gilles Tredan}{laas}
  \end{icmlauthorlist}
  \icmlaffiliation{inria}{Université de Rennes, Inria, CNRS/IRISA (Rennes, France)}
  \icmlaffiliation{laas}{LAAS, CNRS, (Toulouse, France)}
  \icmlaffiliation{peren}{PEReN (Paris, France)}

  \icmlcorrespondingauthor{Gurvan Richardeau}{gurvan.richardeau@peren.gouv.fr}

  \icmlkeywords{Machine Learning, ICML}

  \vskip 0.3in
]

\printAffiliationsAndNotice{} %

\begin{abstract}

Literature reveals that a Large Language Model's (LLM) behavior is not only conditioned by its original weights but also its instance-level parameters, such as instructional prompt, sampling configuration or quantization. A model that generates safe outputs under one configuration may produce toxic content under another.
However, current LLM identification techniques (such as fingerprinting) focus on intellectual property protection, and their design favors robustness to changes in these instance-level parameters. 
This poses a critical challenge for AI regulation in which compliance assessments target actual deployed behaviors, not model provenance. In this paper, we introduce instance-level fingerprinting, a regulator-oriented paradigm that distinguishes configurations of the same LLM. Our method, \scheme, exploits biases in generated binary random sequences. On a benchmark of 237 model instances, it identifies the instance with \nAbsClosedThirtyTwo{} accuracy from 32 queries to the target, against at most \nAbsLLMmapBest{} for the adapted LLMmap baseline. Recognizing that a target is unseen, also called open-set challenge, is much harder. When 10\% of instances are unseen, \scheme realises a recall of \nAbsRecCfifty{} on the unseen instances, at a precision of \nAbsPrecCfifty{}, i.e.\ more than twice better than random. Closed-set instance fingerprinting is thus practical with a few dozen queries, whereas the open-set setting remains an open problem. Code available at \footnote{\url{https://github.com/GurvanR/FLIPS-LLM-Instance-Fingerprinting}}.

\end{abstract}

\vspace{.15cm}
\section{Introduction}
\vspace{.15cm}
Identifying Large Language Models (LLMs) in a black-box setting has recently gained significant traction.
LLM Fingerprinting aims to determine whether a \textit{\reference model} and a \textit{target model} are identical.
It consists of extracting a signature (fingerprint) from the \reference model (\textit{\extraction}) and comparing it with one from the target model (\textit{verification}) \citep{shao2025sok}. %

But what defines identical models? Prior work \citep{pasquini2025llmmapfingerprintinglargelanguage, jin2024proflingo, yang2024fingerprint, gubri2024trap, zhu2025auditing} focuses on Intellectual Property Protection (IPP) scenarios in which the target model is suspected to be a (possibly unauthorized) copy or derivative of the \reference model. In that setting, a \textit{model} is typically defined by the original weights i.e., resulting from a training procedure (since this constitutes the primary cost),
and close derivatives (e.g., fine-tuning, system prompts, quantizations) are \emph{not} considered as distinct models.
In this work, we shift from this IPP scenario to consider regulatory scenarios in which the compliance of an LLM's behavior is evaluated. 
This is motivated by growing legislation surrounding AI governance, most notably the EU AI Act\footnote{\scriptsize \url{https://eur-lex.europa.eu/legal-content/EN/TXT/HTML/?uri=OJ:L_202401689}}, and more recently California's new AI transparency law, SB 53\footnote{ \scriptsize \url{https://legiscan.com/CA/text/SB53/id/3271094}}.

\begin{figure}[t]
    \centering
    \includegraphics[width=0.9\linewidth]{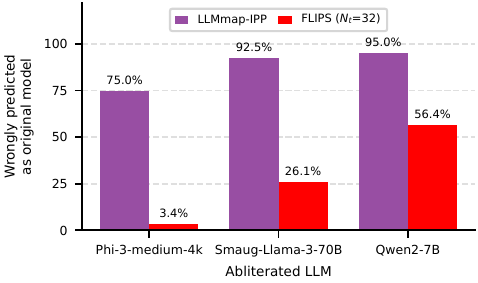}
    \caption{We evaluate the ability of the state-of-the-art IPP fingerprinting scheme LLMmap taken off the shelf \citep{pasquini2025llmmapfingerprintinglargelanguage} and \scheme (ours) to distinguish an "abliterated" (uncensored) instance from its original safe counterpart. LLMmap generally identifies the unsafe abliterated version as the original model, which is the primary goal of the method: being robust to alteration of the original model. \scheme attributes a smaller share of its predictions to the original model, highlighting that IPP-oriented fingerprinting is insufficient for regulators' need, as it fails to detect behavioral shifts that compromise audit compliance.}
    \label{fig:showcase}
\end{figure}

This requires sensitivity to factors beyond original weights, which the literature indicates to be instance-level parameters such as system prompts, sampling configurations, quantization or fine-tuning. We then introduce \textit{Instance-level Fingerprinting} (IF) that targets instances defined as a model along with its precise weights and configuration. Such a fingerprinting method allows regulators to efficiently identify known instances served by any provider, avoiding costly full audits. Critically, IPP-oriented methods deliberately ignore instance-level parameters to achieve robustness, rendering them blind to behavioral modifications. As shown in Figure~\ref{fig:showcase}, the state-of-the-art IPP method LLMmap conflates abliterated\footnote{Abliteration is a weight-editing method used to disable learned safety or refusal behaviors \citep{arditi2024refusal}.} models with their original counterparts. Our method, \scheme, explicitly designed to be sensitive to such changes, makes this confusion less often.

\paragraph{Parameters Affecting LLM's Behavior} 
Literature shows that system prompts can alter model behavior, \citet{neumann2025position} shows that system prompts such as "you are a Christian" lead to very different definitions of heterosexuality, while \citet{mou2024sg, cui2024or} show that a safe-oriented system prompt can effectively produce a safer version of the model. \citet{bianchi2023safety} shows that fine-tuning on toxic code induces generalized toxic behavior. Quantization can increase stereotype biases \citep{marcuzzi2025quantization} or harmfulness \citep{kharinaev2025investigating, marchisio2024does} and can increase vulnerabilities \citep{belkhiter2024harmlevelbench}. Finally, temperature also matters \citep{cecere2025monte}, where changing from 0.6 to 1.0 can affect benchmark scores as much as 10\% \citep{du2025optimizing}. 

\paragraph{\scheme Rationale} Such a sensitive fingerprinting scheme must be triggered by subtle instance-level parameters. Among them, the temperature poses a particular challenge. Recent work has investigated LLMs' ability to generate random sequences when prompted for it \citep{harrison2024comparison,hopkins2023can,van2024random}. \citet{coronado2025deterministic} and \citet{van2024random} notably show that temperature can have a great influence on output distribution. More generally, most models demonstrate poor randomness generation capabilities. Model failures on specific tasks have proven effective for model identification, from accuracy differences to classification failure patterns on simple images \citep{maho2023model}. In that light, we choose to leverage the discriminative power of random generation biases for LLM fingerprinting. We focus on a simple task: generating binary sequences of two tokens. This minimal sample space facilitates the isolation of instance-specific biases from limited samples and allows capturing those biases reliably using the well-established NIST cryptographic test suite \citep{bassham2010statistical}. This novel approach offers a key advantage over neural network-based fingerprinting. LLMmap relies on pretrained text embedders to process semantic sequences, resulting in opaque decision boundaries \citep{kim2020transparency}. In contrast, the NIST test suite assigns specific deviations from randomness to particular LLMs.

This paper makes the following contributions: (1) Introduce and motivate Instance-level fingerprinting and exhibit limits of current approaches for regulators' needs. 
(2) Present \scheme, a query-efficient method for stealthy extraction and verification of instances (summarized in Figure \ref{fig:flips_schema}). (3) Validate the approach against a comprehensive benchmark of 237 instances derived from 25 LLMs, reaching a closed-set accuracy of \nAbsClosedThirtyTwo{} with 32 verification queries and, in the open-set setting, detecting unseen instances more than twice better than random, with a threshold fixed without any oracle.

\section{Problem Setting}
\label{sec: Problem Setting}

We formalize the instance-fingerprinting problem from the perspective of a regulator, though broader applications could be found, such as for platform operators and other stakeholders seeking to verify model deployments.  
The regulator wants to ensure that the model behaves identically to how it behaved when audited; in that light, an \textit{ instance} is defined by its precise weights (quantization, fine-tuning included), its instructional framing (e.g., system prompt), and the generation hyperparameters (e.g., temperature, top-$k$). 
We assume an honest instance behavior, that does not actively seeks to detect fingerprinting and manipulate its outcome. Let $\mathcal{M}$ be the set of all possible instances.

\paragraph{Two-Stage Protocol}
As traditionally done in model fingerprinting approaches, instance-fingerprinting operates in two stages:

\begin{enumerate}
\item \textbf{Extraction} The regulator constructs a database of reference instances, denoted $E \subset \mathcal{M}$. For each reference instance $m_r \in E$, the regulator extracts a fingerprint via $\xi: m \mapsto \mathbf{F}$, that maps an instance to a fingerprint. In addition, the regulator performs domain-specific evaluations (e.g., toxicity benchmarks, capability assessments) to characterize the instance properties.

\item \textbf{Verification} When encountering a deployed target instance $m_t$, the regulator checks if it matches any reference instance in $E$. If matched, the regulator retrieves the associated evaluation results, avoiding re-running it; otherwise, it adds the target instance to $E$ using the extraction procedure.
\end{enumerate}

\paragraph{Operational Requirements}
The fingerprinting system must satisfy the following constraints:

\begin{itemize}
\item \textbf{Strict Black-Box Access:} The regulator possesses only the access granted by the model provider for both extraction and verification. So in general, we assume no access to weights or logits, and interactions are limited to sending a text query and receiving
a generated text response.

\item \textbf{Open-Set Capability:} Let $D$ be the decision function that maps a fingerprint to $E \cup \{\textit{Unseen}\}$. Given a target instance $m_t$, $D$ must either match it to its reference in $E$ ($D(\xi(m_t)) = m_t$ when $m_t \in E$) or correctly recognize it as unseen ($D(\xi(m_t)) = \textit{Unseen}$ when $m_t \notin E$).

\item \textbf{Query Efficiency:} We seek to minimize the total number of requests. We distinguish between two primary query budgets: (1) \textit{Extraction queries} ($N_r$), used to build $E$, where a larger budget can be tolerated to ensure fingerprint robustness, provided the cost remains substantially lower than a full evaluation; and (2) \textit{Verification queries} ($N_t$), used for matching target instances, which must be strictly minimized to facilitate routine verification. Beyond operational costs, high query efficiency also helps evade detection by model providers who may monitor for anomalous usage \citep{zhao2025}.
\end{itemize}

\paragraph{Performance Metric}
Due to the stochastic nature of LLM generation\footnote{Stochasticity arises from sampling procedures (e.g., top-$k$, nucleus sampling) and hardware non-determinism \citep{atil2024non}.}, we model $\xi(m)$ as a random variable.
We define the primary performance metric as the expected accuracy under this open-set criterion:
\begin{align}\label{eq: final accuracy def}
    \textit{acc}(D)
    &= \mathbb{E}_{m\in\mathcal{M}} \left[\mathbb{P}\!\left(D(\xi(m)) = \tilde{m} \,\middle|\, m \right)\right], \\
    \text{with} \quad \tilde{m} &= \begin{cases} m & \text{if } m \in E, \\ \textit{Unseen} & \text{if } m \notin E. \end{cases} \notag
\end{align}
\section{\scheme Fingerprinting Scheme}
\label{sec:flips_scheme}

\begin{figure}[t]
    \centering
    \includegraphics[width=0.95\linewidth]{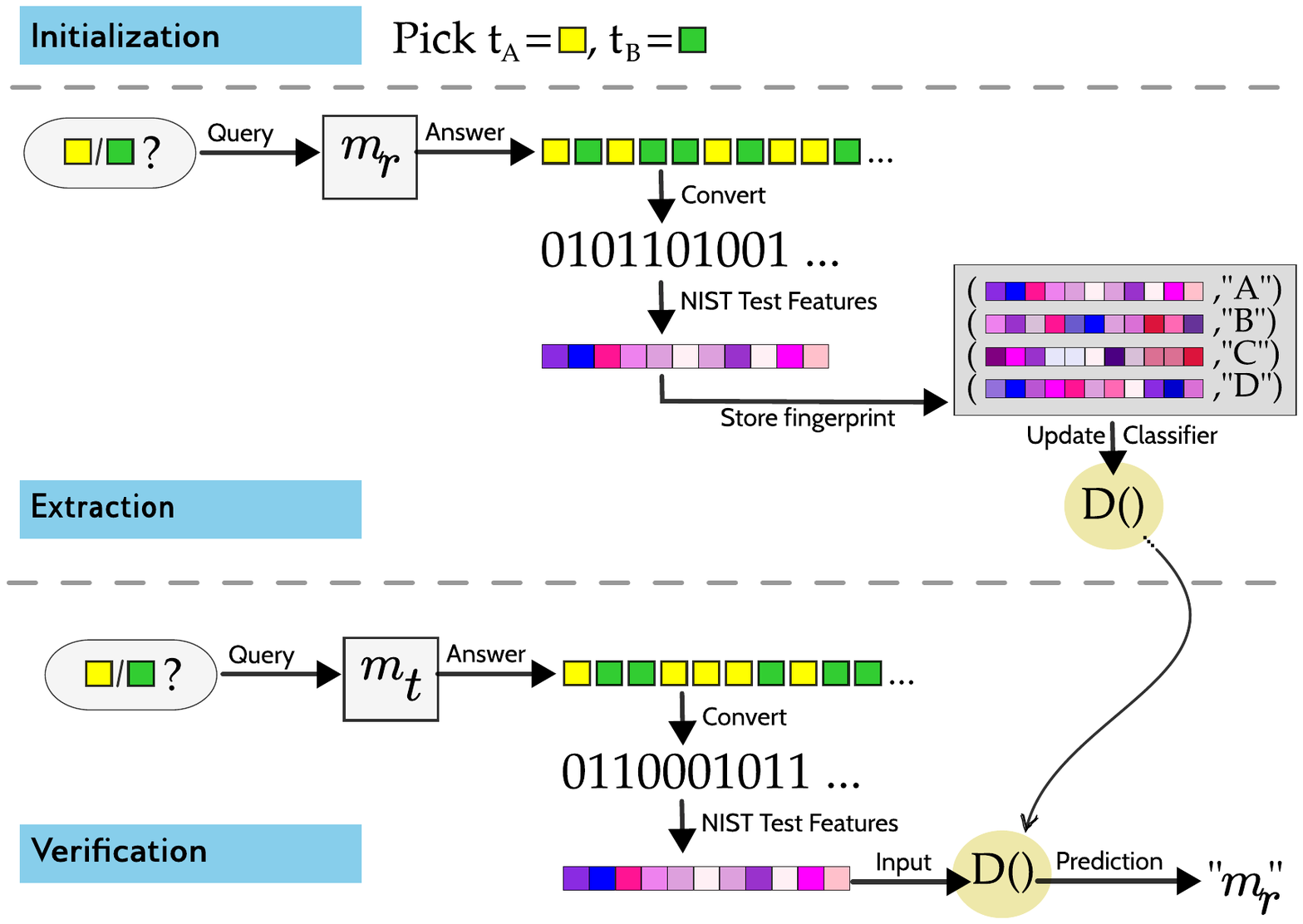}
    \caption{%
    The \scheme fingerprinting method. (1) Two tokens $t_A$ and $t_B$ are selected to query the \textit{\reference} instance $m_r$. (2) Extraction: $m_r$ generates random binary sequences that are then converted into bit sequences to pass the NIST randomness test suite. The resulting test statistics, with complementary statistics of the bit sequences, serve as features to an XGBoost classifier. (3) Verification: the \textit{target} instance $m_t$ is asked to generate random sequences (with the same $t_A, t_B$), that are fed to the trained classifier to predict either a reference instance or \textit{Unseen} label.}
    \label{fig:flips_schema}
\end{figure}

This section describes how we implement \scheme in this instance-level fingerprinting. An overview is provided in Figure \ref{fig:flips_schema}.

\paragraph{Querying Strategy} \scheme employs a fixed prompt template $q_0$ (formalized in Appendix \ref{appendix: Prompt q_0}) that instructs the instance to generate a random binary sequence. The baseline prompt requests symbols \texttt{0} and \texttt{1} (yielding sequences such as \texttt{011100101...}). Alternative formulations substitute different token pairs; for instance, \texttt{car} and \texttt{sun} produce sequences like \texttt{carsunsunsuncar...}, which are subsequently mapped to binary form (e.g., \texttt{car} $\to$ \texttt{0}, \texttt{sun} $\to$ \texttt{1}). Token pairs are formed from the alphanumeric tokens shared by all reference models; we denote $\mathbf{T}$ the resulting set of token pairs\footnote{Restricting to alphanumeric tokens makes this set likely to generalize to models for which we cannot inspect tokenizers directly.}.

A deployment fixes one token pair $\mathcal{S} = (t_A, t_B) \in \mathbf{T}$; extraction and verification use the same pair, and each query yields one sample $o \sim m(q_0(\mathcal{S}))$ with output in $\mathcal{O} = \{0,1\}^* = \bigcup_{\ell \in \mathbb{N}} \{0,1\}^\ell$. The accuracy depends on the token pair (Appendix~\ref{appendix: Token Pair examples}).
\paragraph{Fingerprint} The \scheme fingerprint relies on a feature map $f : \mathcal{O} \longrightarrow \mathbb{R}^d$ corresponding to the $d=57$ statistics of the bit sequence: the $31$ test statistics of the NIST suite, the sequence length and $25$ complementary statistics (Appendix \ref{appendix: nist-tests}), defined as: 
\begin{align} \label{eq:embedding_function}
    \xi_{N}^{\mathcal{S}}: \,  \mathcal{M} &\longrightarrow \mathbb{R}^{d \times N} \notag \\
           m &\longmapsto \mathbf{F} = \left(f(o_i)\right)_{1 \leq i \leq N}
\end{align}
where $o_{i} \sim m(q_0(\mathcal{S}))$ i.i.d.\ for $i \in \{1,\dots,N\}$. The budget $N$ is generic: it is $N_r$ at extraction and $N_t$ at verification.

\paragraph{Decision function} We cast the fingerprinting task as a multi-class classification problem where $D$ is built from XGBoost classifiers trained on the reference-set fingerprints $\left(\xi_{N_r}(m)\right)_{m \in E}$.
$D$ is defined as
\begin{align} \label{eq:scoring_function}
    D : \mathbb{R}^{d \times N_t} &\rightarrow E \cup \{\textit{Unseen}\}, \\
    \mathbf{F}_{m_t} &\mapsto \argmax_{k} \hat{y}_k, \quad
    \hat{y}_k = \mathbb{P}( m_k \mid \mathbf{F}_{m_t}), \; k \in [|E|]. \notag
\end{align}
Given a fingerprint, an XGBoost classifier trained on all the instances of $E$ computes $\hat{\mathbf{y}}$, a probability distribution over instances in $E$. The predicted instance is determined by $\argmax_{k} \hat{y}_k$. $D$ combines two parts, described next: an \emph{instance classifier}, which names an instance of $E$, and an \emph{Unseen detector}, which decides whether $D$ returns \textit{Unseen} instead.

\paragraph{Instance classifier} Instances of different LLMs are easier to tell apart than instances of one LLM (Section~\ref{s:evaluation}). The instance classifier therefore works in two steps (Algorithm~\ref{alg:TwoStage}, Appendix~\ref{appendix:evaluation-procedure}). First, the classifier of Eq.~\ref{eq:scoring_function} selects the LLM of the target: the LLM of the instance it predicts. Second, a classifier trained on the instances of that LLM only decides among them.

\paragraph{Unseen detector} The Unseen detector is a logistic regression on statistics of the output of the instance classifier for the target (Appendix~\ref{appendix:evaluation-procedure}): it computes an Unseen probability $q$, and the target is labelled \textit{Unseen} iff $q \geq \tau$, where the threshold $\tau$ is fixed at extraction. In a realistic setup, no \textit{Unseen} instance is available at that time, neither to train the detector nor to set $\tau$ (no oracle assumption). We therefore simulate them: instances of $E$ are held out in turn from the training, and they then play the role of \textit{Unseen} instances (Algorithm~\ref{alg:BuildDetector}). The detector is trained on these simulated \textit{Unseen} instances, and $\tau$ is set so that they are flagged $50\%$ of the time. This $50\%$ is thus a target on simulated \textit{Unseen} instances; the share of real \textit{Unseen} instances that is flagged is a result (Section~\ref{s:evaluation}). The threshold so obtained is our \emph{operating point}.
\paragraph{Extraction Phase} The \extraction phase constructs the decision function $D$ through the following steps:

\begin{enumerate}
    \item \textbf{Collecting Data.} For each instance in $E$, query the instance $N_r$ times with prompt $q_0(\mathcal{S})$.

    \item \textbf{Feature Extraction.} Convert each output sequence to a binary sequence $o \in \{0,1\}^*$ and compute the features $\mathbf{x} = f(o) \in \mathbb{R}^d$.

    \item \textbf{Classifier Training.} Train the instance classifier on the extracted features $\{\mathbf{x}_{m,i}\}_{m \in E, i \in \{1, \dots, N_r\}}$; for the open-set setting, build the \textit{Unseen} detector from the same features (Algorithm~\ref{alg:BuildDetector}).%
\end{enumerate}

\paragraph{Verification Phase} Given a target instance $m_t$ and the trained decision function $D$, the verification procedure queries $m_t$ $N_t$ times with $q_0(\mathcal{S})$ and computes the final prediction $D(\xi^{\mathcal{S}}_{N_t}(m_t))$.

\paragraph{Remark} Because the XGBoost classifiers are trained on single answers, and their outputs on the $N_t$ answers of a target are aggregated at verification (Appendix~\ref{appendix:evaluation-procedure}), the verification budget $N_t$ is chosen at deployment time, trading accuracy for queries, without retraining.

\section{\scheme Discriminative Power}\label{s:discpow}
Before evaluating \scheme empirically, we examine the origin of its discriminative power by exploring the randomness of the generated sequences.\footnote{This section is purely explanatory: it leverages white-box log-probabilities to explain \emph{why} \scheme works, whereas the \scheme pipeline itself relies only on black-box text outputs.}

To generate high-quality random sequences conditioned on a token pair $(t_A,t_B)$, an LLM must assign near-equal probability to both tokens, allowing the pseudo-random sampling algorithm to produce unbiased output. We therefore investigate the log-probabilities assigned to both tokens at each generation step.

\paragraph{Token-Level Analysis}
\begin{figure}[h!]
    \centering
    \includegraphics[width=0.48\textwidth]{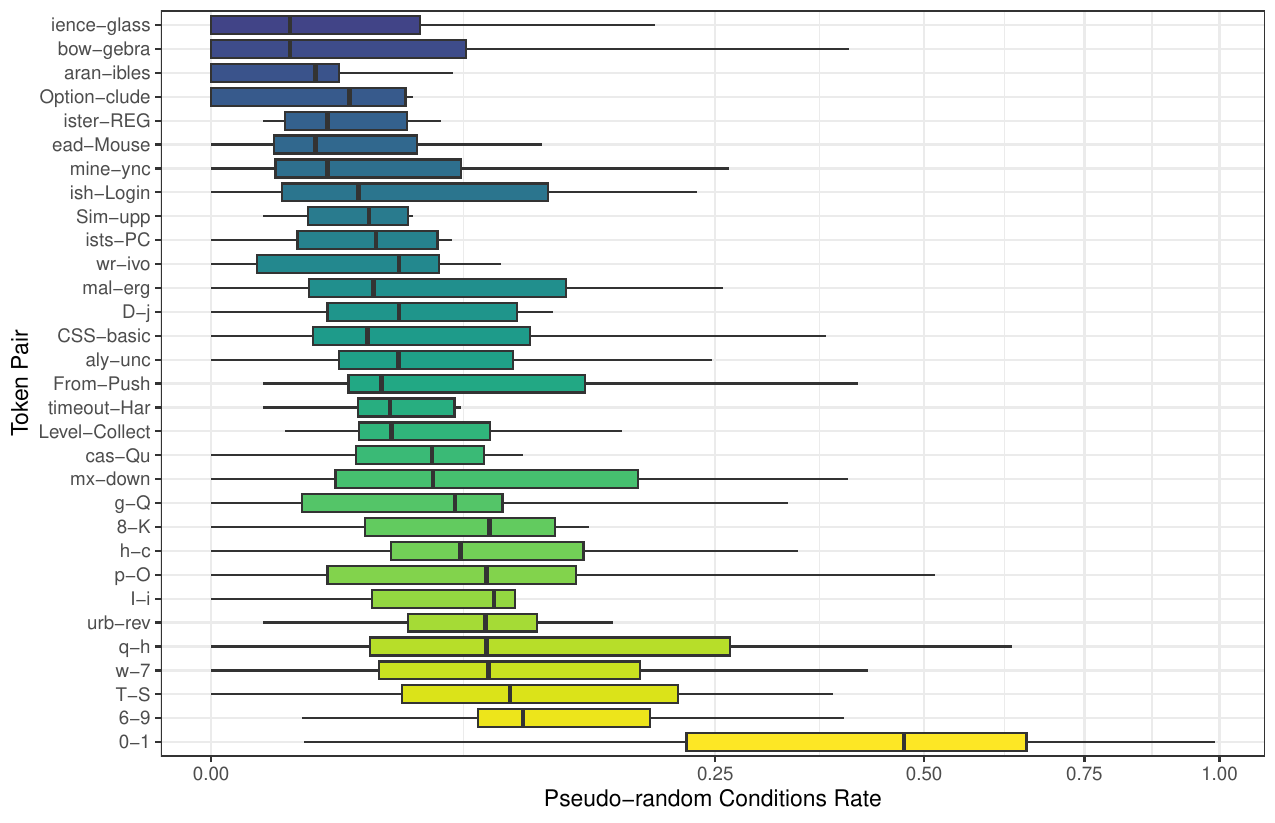}
    \caption{\textbf{Calculation of the Pseudo-random Conditions Rate.} For each token pair $(t_A, t_B)$, we evaluate every generated token across all sequences and models by comparing their log-probabilities. A binary score is assigned based on whether the log-probability of $t_A$ is significantly higher than $t_B$ or if they are relatively similar. This score is then averaged, yielding the Pseudo-random Conditions Rate.}
    \label{fig:logprobs_gen}
\end{figure}

To further investigate this observation, we select a set of arbitrary token pairs and models, and explore the log-probabilities returned during sequence generation. For each output token generated, we seek to distinguish between two extreme situations: either $p(t_A) \approx p(t_B)$, in which case the actual token choice is left to the sampler, or $p(t_A)\gg p(t_B)$ (or vice-versa), in which case the token choice is determined before the sampler.%

Formally, to capture these pseudo-random conditions, we define: $\beta
= \mathds{1}_{\left\{\left|\log({p(t_A)})-\log({p(t_B)})\right| < \log 2\right\}}$. Figure~\ref{fig:logprobs_gen} shows this metric for  various token pairs, aggregated across all generated sequences and models. %
The $\beta$ rate is generally below 25\%, indicating that LLMs typically assign high confidence to one token over the other. This reveals strong internal blueprints influencing the construction of such sequences. 
The existence of this blueprint introduces instance-specific biases: since the sampler rarely determines token choice, decisions arise from the complex interplay of weights, attention mechanisms, context, and hyperparameters. Each component (such as temperature and system prompt) contributes distinctively to the resulting sequences. Moreover, since this blueprint is likely arbitrary rather than principled (as are pseudo-random algorithms), it varies across instances. Those mechanisms render the output highly sensitive to the alteration of even a single contributor.
The expression of this instance-specific blueprint translates into specific biases that explain \scheme' high discriminative power.
The \texttt{0-1} token pair is a notable exception, discussed below.

\begin{figure}[h!]
    \centering
    \includegraphics[width=0.48\textwidth]{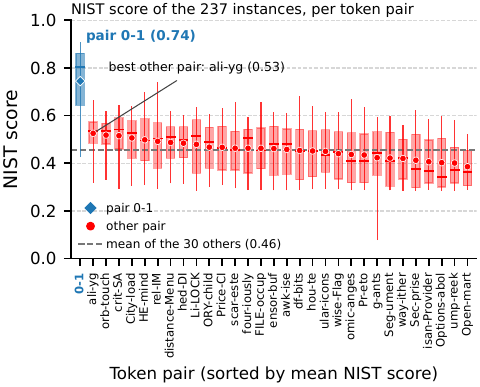}
    \caption{{Randomness quality of the token pair \texttt{\small 0-1} against each of the other token pairs: NIST score (mean over the $237$ instances of the average share of the sequences that pass the NIST tests) of each of the \nNPairs{} pairs (box: quartiles, whiskers: 5th to 95th percentile, dot: mean).}}
    \label{fig:zero_one_vs_pairs}
\end{figure}

\paragraph{Randomness Quality and Fingerprinting} 
Figure~\ref{fig:zero_one_vs_pairs} shows that the sequences exhibit poor randomness quality. 
The \texttt{\small 0-1} token pair shows particularly higher $\beta$ score and randomness quality: its mean NIST score is \nFourNistZeroOne{}, against \nFourNistOthers{} on average for the other pairs, and it is the highest of all pairs for \nFourShareBeatsAll{} of the instances. Unlike arbitrary token pairs such as \texttt{\small scar-este}, binary sequences (0s and 1s) are very common in training corpora. Hypothetically, this prevalence enables models to better memorize distributional patterns of binary sequences, reducing instance-specific biases and improving randomness; yet its fingerprinting recall at $N_t=32$ (the share of its verification samples identified as that instance, closed-set) is ordinary (\nFourRecallZeroOne{}, against \nFourRecallOthers{} for the other pairs).

\paragraph{Feature Importance}
To illustrate the statistical features leveraged by \scheme, we examine their importance (XGBoost gain) in the instance classifier: the top $25$ features are in Figure~\ref{fig:feature_importance} and all statistics are listed in Table~\ref{tab:fullNistTest} (both in Appendix~\ref{appendix: nist-tests}; NIST test descriptions based on \citealt{bassham2010statistical}). No single statistic dominates: the largest share of the gain carried by one feature is \nImpTop{}. The $31$ NIST statistics carry \nImpShareNist{} of the gain, the statistics of the dependence on past bits \nImpSharePast{}, the opening-bit statistics \nImpShareOpen{} and the sequence length \nImpShareLen{}. The contribution of the sequence length is measured separately, by truncating all sequences to a fixed length (Appendix~\ref{appendix:seqlen_truncation}).

\section{Experimental Evaluation}
\label{s:evaluation}
Having characterized the source of \scheme' discriminative power, we now quantify its operational performance. 

\subsection{Experimental Setup}

\paragraph{Benchmark}
We create an extensive benchmark comprising 237 instances derived from 25 open-weight LLMs. To construct it, we produce eight variations per LLM: four temperature settings ($0.4, 0.6, 0.8, 1.0$) without system prompt, and four distinct system prompts at temperature $1.0$. In addition, for 4/25 models we include two quantization variants (\texttt{int4}, \texttt{fp8}), each in four of these settings (temperature $0.6$ and $1.0$ without system prompt, \texttt{sp1} and \texttt{sp2}). Finally, abliterated instances are included when available (5/25 models). 
We adopt $0.2$ temperature increments, as these intervals have been shown to significantly influence LLM benchmark performance \citep{du2025optimizing}. The system prompts, detailed in Appendix~\ref{appendix:system_prompts}, were selected for their documented impact on model behavior: \texttt{sp1} is known to enhance model safety \citep{cui2024or}, while \texttt{sp2} has been shown to increase harmfulness \citep{mou2024sg}. Finally, \texttt{sp3} and \texttt{sp4} instruct the model to adopt Conservative and Liberal personas, respectively, alignments proven to significantly shift model opinions \citep{neumann2025position}. 
The resulting $(25$LLM $\times 8$variations$)+ (4$LLM$ \times 2\times 4$quantization variations)$ $$+5$abliterated$=237$ instances constitute the benchmark of instances that \scheme has to differentiate.

\paragraph{Inference Setup}\label{appendix: Generation Parameters}
All open-weight instances are queried via vLLM \citep{kwon2023efficient} on A100 80GB GPUs, with \texttt{bfloat16} base weights. The context window is $2048$ tokens, with at most $500$ generated tokens per query. Sampling uses the default Hugging Face configuration top-$k=50$, top-$p=1.0$ and repetition penalty $1.0$, on top of the temperature and system-prompt variations described above. The \texttt{int4} variants use vLLM's \texttt{bitsandbytes} (NF4) backend; the \texttt{fp8} variants use vLLM's native fp8 quantization. We collect $500$ sequences per instance per token pair. To mitigate truncated outputs, any generation yielding fewer than $100$ bits after token-pair extraction (scanning the output and mapping each $t_A$/$t_B$ to bit $0$/$1$) is re-queried up to $5$ times; if all retries fail, the sequence is discarded and logged, with per-instance discard rates reported in Figure~\ref{fig:proper-answers}.

\paragraph{Decision function}
We construct the decision function $D$ by querying each of the  237 instances $N_r=300$ times (fewer for the few instances that have fewer valid sequences left); the test set holds $128$ sequences per instance on average. A verification sample consists of $N_t$ distinct test sequences of one instance, with $N_t$ from $1$ to $64$: the test sequences of an instance are cut into disjoint samples, about $128/N_t$ per instance. Unless stated otherwise, the results are given at $N_t=32$.
As discussed in Section~\ref{sec: Problem Setting}, open-set is a key requirement that we evaluate as the main setup, in addition to a closed-set environment. We employ repeated random splits for both closed-set and open-set evaluations. In the open-set setup, instances are partitioned into disjoint \emph{Known} and \emph{Unseen} subsets, with $10\%$ held out as \emph{Unseen} ($23$ of $237$ instances, drawn uniformly at random at each of $5$ draws); the reference set contains only \emph{Known} samples, while the target set includes both.  In the closed-set setup, all instances appear in both reference and target sets, over $5$ random splits of the sequences. Evaluation details are in Appendix~\ref{appendix:evaluation-procedure}. Appendix~\ref{appendix:clf_choice} justifies the XGBoost choice against standard classifier baselines.

\paragraph{\scheme Token Pairs} To study the stability of performance across different choices of token pairs, we run both evaluations on each of \nNPairs{} pairs: $30$ sampled from $\mathbf{T}$ (see Appendix \ref{appendix: Token Pair examples} for examples) and the specific \texttt{0-1} pair whose distinctive randomness behavior was analyzed in Section~\ref{s:discpow}. All values are means over the pair means, with standard deviations across pairs for the bands of the figures.

\paragraph{LLMmap IF Adaptation} We adapted LLMmap to instance-level fingerprinting following the official implementation at {\small \url{https://github.com/pasquini-dario/LLMmap}}, specifically the ``Train your own model'' section. All classifier pipeline hyperparameters were kept at their default values, with $N_r=40$ training samples per instance. In experiments varying the number of queries $N_t$ (Figure~\ref{fig:FLiPS_vs_LLMmap_bs}), we selected the first $N_t$ queries from this fixed set in sequential order, as recommended in the paper \citep{pasquini2025llmmapfingerprintinglargelanguage}.

\subsection{General Performance}

\begin{figure}[t]
    \centering
    \includegraphics[width=0.45\textwidth]{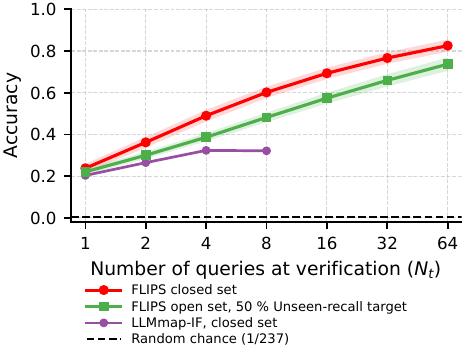}
    \caption{Accuracy as a function of verification queries ($N_t$) for \scheme and LLMmap (adapted for IF). 
    Closed-set: instance classifier of Section~\ref{sec:flips_scheme} (mean over \nNPairs{} token pairs, band: standard deviation across pairs). Open-set (Eq.~\ref{eq: final accuracy def}, all token pairs and draws pooled): our method at its operating point (Section~\ref{sec:flips_scheme}; the $50\%$ Unseen-recall target is set on simulated \emph{Unseen} instances). LLMmap-IF: closed-set, up to $N_t=8$.}
    \label{fig:FLiPS_vs_LLMmap_bs}
\end{figure}

Figure~\ref{fig:FLiPS_vs_LLMmap_bs} presents the overall classification accuracy as a function of $N_t$, the number of queries in the verification stage. 
In the closed-set setting, where all target instances are known, \scheme with its instance classifier (Section~\ref{sec:flips_scheme}) reaches \nClosedThirtyTwo{} accuracy (\nClosedEight{} at $N_t=8$, and \nClosedSixtyFour{} at $N_t=64$, where the curve is still rising). In the open-set setting, the accuracy at our operating point (Section~\ref{sec:flips_scheme}) is \nCfiftyAccThirtyTwo{}.

In comparison, LLMmap, in closed-set, despite being specifically trained on these instances, offers substantially less performance, reaching at best \nLLMmapBest{} accuracy (at $N_t=\nLLMmapBestNt$). LLMmap performance is stable from 4 (\nLLMmapFour{}) to 8 queries, suggesting that the number of verification samples does not constitute the performance bottleneck in this setting; we therefore do not run it beyond $N_t=8$, where it has plateaued and each run is computationally expensive. 
We do not evaluate LLMmap in the open-set setting: its weak closed-set result would only degrade further, and retraining it across the open-set cross-folds costs roughly $20\times$ \scheme' XGBoost pipeline. These results highlight that LLMmap, despite its recognized performance in model-fingerprinting tasks, is not a viable instance-fingerprinting solution.

\begin{figure}[t]
    \centering
    \includegraphics[width=0.9\linewidth]{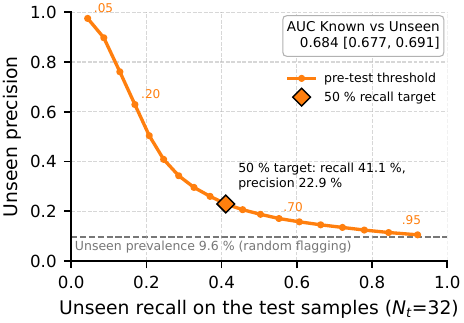}
    \caption{Unseen precision against Unseen recall of our method at $N_t=32$ (mean over the \nNPairs{} token pairs). Each point is a threshold fixed before the test to flag a target share ($5\%$ to $95\%$) of the samples of simulated \emph{Unseen} instances; it is placed at the recall it realises on the test samples. The diamond is the $50\%$ target, our operating point. Text: AUC of the Unseen probability on the \emph{Known}/\emph{Unseen} task.}
    \label{fig:openset_pr_tradeoff}
\end{figure}

\paragraph{Open-set Precision/Recall Trade-off.} The open-set headline number above hides a trade-off between identifying \emph{Known} instances and rejecting \emph{Unseen} ones, governed by the threshold $\tau$ (Section~\ref{sec:flips_scheme}). 
Figure~\ref{fig:openset_pr_tradeoff} shows its detection side: the Unseen precision against the Unseen recall as the threshold varies (each point is one threshold, see the caption). The Unseen probability separates \emph{Known} from \emph{Unseen} samples with an AUC of \nAucThirtyTwo{}. The performance on \emph{Unseen} instances is not high: open-set detection is challenging: the Unseen probabilities of \emph{Known} and \emph{Unseen} samples overlap strongly (Figure~\ref{fig:openset_thresholds}), and Section~\ref{s:detailed} examines why. At our operating point (set without oracle), \scheme flags \nRecCfiftyThirtyTwo{} of the real \emph{Unseen} samples with a precision of \nPrecCfiftyThirtyTwo{}, which is \nPrecLiftCfifty{}$\times$ better than random flagging (\nUnseenPrev{} of the target samples are \emph{Unseen}).

\subsection{Detailed Performance Results}\label{s:detailed}

\begin{figure*}[h]
    \centering
    \includegraphics[width=0.9\textwidth]{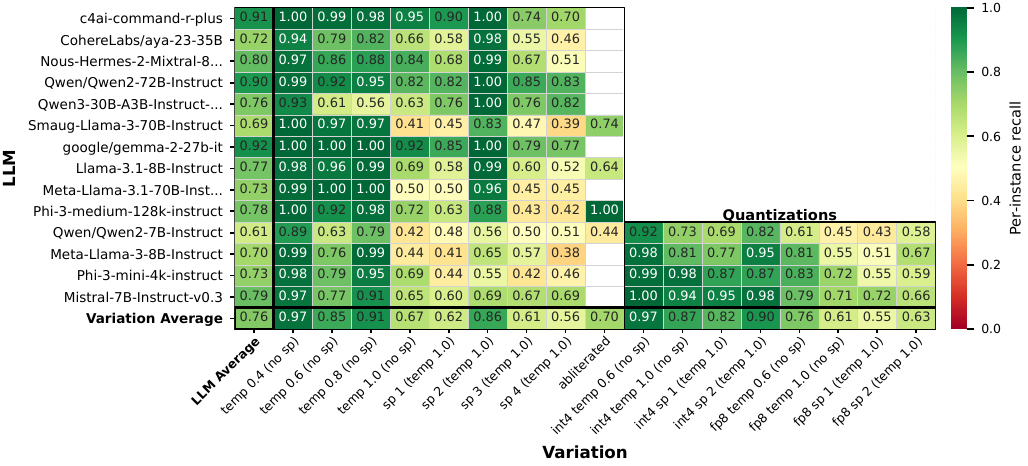}
    \caption{\scheme per-instance recall in the closed-set setting with $N_t=32$ (shortlist of representative instances, full heatmap in Appendix~\ref{appendix:full_results}, Figure~\ref{fig:FullResultsHeatmap}; open-set counterpart in Figure~\ref{fig:FullResultsHeatmapOpenSet}). Each tile is the recall on an instance (base LLM plus variation), with the instance classifier, mean over the \nNPairs{} token pairs; \texttt{temp} and \texttt{sp} refer to temperature and system prompt, \texttt{int4} and \texttt{fp8} to the quantizations. }
    \label{fig:ShortResultsHeatmap}
\end{figure*}
Figure~\ref{fig:ShortResultsHeatmap} details per-instance recall for a subset of instances (all $237$ in Figure~\ref{fig:FullResultsHeatmap}). {Averaged over the $237$ instances, the recall is \nHeatGrand{}; it differs strongly between base LLMs, from \nHeatLlmLow{} for \texttt{\small \nHeatLlmLowName{}} to \nHeatLlmHigh{} for \texttt{\small \nHeatLlmHighName{}}, and between variations (variation averages from \nHeatVarLow{} to \nHeatVarHigh{} among the unquantized variations).} %
The harmful system prompt \texttt{sp2} (see Appendix~\ref{appendix:system_prompts}), is {the most identifiable system prompt (\nHeatSpTwo{} on average, against \nHeatSpThree{} and \nHeatSpFour{} for \texttt{sp3} and \texttt{sp4}), and it is particularly salient for \texttt{\small Qwen2-72B-Instruct} and \texttt{\small Qwen3-30B-A3B-Instruct}, achieving \nHeatQwenSeventyTwoSpTwo{} and \nHeatQwenThirtySpTwo{} recall respectively on this variant}. Higher temperatures translate to lower fingerprinting: from \nHeatTempZeroFour{} at temperature 0.4 to \nHeatTempOneZero{} at temperature 1.0 (without system prompt). This trend is expected, as higher temperatures increase output entropy. Regarding quantization, the \texttt{fp8} variants are the hardest to identify (\nHeatFpEight{} at temperature $1.0$ without system prompt, against \nHeatIntFour{} for \texttt{int4}), and the abliterated instances reach \nHeatAbli{}. %

\paragraph{Confusion Analysis} 
Figure~\ref{fig:Confusion_orig_closed} %
(Appendix~\ref{appendix:full_results}) shows how often \scheme misses each attribute of an instance in the closed-set setting, relative to an instance drawn at random: it misses the system prompt \nAttrRelSP{} as often as random, the quantization \nAttrRelQuant{}, the temperature \nAttrRelT{} and the base LLM \nAttrRelModel{}. The hardest distinctions are therefore between instances of the same LLM that differ in their system prompt or their quantization. Unseen instances are confused in a similar way: for \nAbsDistOne{} of the Unseen samples, the Known instance chosen by the instance classifier differs from the Unseen one in a single attribute. This is why they are hard to detect: when every Unseen instance is at an attribute distance of at least $2$ from every Known one, the AUC of the Unseen detector rises from \fx{2}{\nDtwAucUsualThirtyTwo} to \fx{2}{\nDtwAucDtwoThirtyTwo} (Appendix~\ref{appendix:distance2}, Figure~\ref{fig:distance2}).%

\section{Discussion}
\label{s:discussion}

\subsection{Regulation Perspective}\label{sec:regulator_discussion}

\paragraph{Procedural Regulation and Technical Standards} Instance-level fingerprinting aligns with emerging AI governance frameworks. The EU AI Act\footnote{\scriptsize \url{https://eur-lex.europa.eu/legal-content/EN/TXT/HTML/?uri=OJ:L_202401689}} (2024) has tasked\footnote{\scriptsize \url{https://ec.europa.eu/transparency/documents-register/detail?ref=C(2025)3871&lang=en}} standardization organizations like CEN and CENELEC\footnote{\scriptsize \url{https://www.cencenelec.eu/european-standardization/european-standards/}} with developing technical standards for AI systems. As \citet{crum2025brussels} notes, rather than prescribing specific technical requirements, the Act "adopts a markedly procedural strategy" that leaves concrete implementation methods to be determined.

\paragraph{Access Regime Under the AI Act} In the AI Act, LLMs are classified as general-purpose AI, and their providers' obligations (Article 53) primarily concern providing documentation. Full access to a model is not granted by default: it can be requested, but has to be proven necessary and proportionate (Articles 91, 92), and is likely to face delays as well as intellectual-property or security concerns (Article 78). While white-box methods remain essential to develop for cases where such access is granted, we think that black-box techniques with minimal assumptions are essential for building evidence in practice. They also enable independent audits by academic researchers and civil society without requiring privileged access.

\paragraph{Post-deployment Versioning Tracking} Beyond the white/black-box question, regulation increasingly weighs continuous monitoring. Article 53 of the AI Act and chapter 25.1 of SB53\footnote{\scriptsize\url{https://leginfo.legislature.ca.gov/faces/billNavClient.xhtml?bill_id=202520260SB53}} explicitly require that providers keep their technical documentation up-to-date: this can in principle be tested through independent verifications. \citet{reuel2024open} point out that a key open question is precisely the tracking of post-deployment model versioning, especially for frequent model updates. \scheme offers a practical implementation that tracks changes over time, registers trusted fingerprints, and flags new versions (yet still challenging). Moreover, its query efficiency (a few dozen verification queries) makes such continuous monitoring operationally feasible even under frequent updates.

Finally, as a more specific example, \citet{sokhansanj2025local} highlights the  regulatory blind spots created by the rise of open-weights LLMs.

\paragraph{License plate analogy} We focus on the fingerprinting of honest instances. We consider \scheme as the equivalent of license plates: an easy way for the regulator to identify an instance. License plates can be forged, but this criminal behavior is considerably more involved and risky. Likewise, active fingerprinting deception is a considerably more challenging endeavor. Despite their fragility, license plates remain a staple of driving regulation technology, easily covering the vast majority of use cases.

\subsection{Capabilities and Limitations of \scheme}
Considering the many aspects at play in the fingerprinting of modern LLMs, we now discuss the most salient ones, for potential \scheme limitations but also future work.
\paragraph{LLMs Accessing Tools} The advent of code execution capabilities in frontier LLMs, in particular with agents, might suggest that they can generate random outputs through computational means rather than relying on their inherent stochastic processes. However, current implementations keep code execution transparent and under user control, and this is likely to persist.

\paragraph{Agent Fingerprinting} In the same vein, the rise of agentic AI systems introduces new challenges for fingerprinting tasks. These systems remain nevertheless fundamentally LLM-based, which will likely make them fingerprintable through conventional text generation queries irrespective of their sophisticated scaffolding or multi-LLM configurations.

\paragraph{LLM Capabilities Expansion} As LLMs advance, they may develop enhanced random generation capabilities without requiring code execution, reducing the discrepancies that \scheme leverages. For instance, recent research has demonstrated that LLMs can internally embed deterministic algorithms within their weights, such as addition and multiplication schemes \citep{kantamneni2025language, maltoni2024arithmetic}. However, this challenge can be addressed through progressively sophisticated randomness tasks: transitioning from binary to Poisson processes, or demanding complex random graph generation \citep{richardeau2025llms}.

\paragraph{Adversarial Query Rerouting} The query-detection vulnerability (a provider recognizing the audit's query pattern and substituting another generator) is an important defense yet shared by all other fingerprinting methods because they rely on specific queries, and remains an open challenge in the literature. The defense available to \scheme is itself general: when the substitute generator is itself fingerprintable, it can simply be enrolled as an additional instance and flagged by the same pipeline. As a concrete example, consider a provider rerouting random-generation queries to a true (quality) PRNG. Such a trick would be easy to flag, since true PRNG sequences are easily detected as such by NIST tests, which is their original purpose.

\section{Related Work}
\label{s:related}

\subsection{Field Clarification}

\paragraph{\textit{Fingerprinting}, not \textit{watermarking}} Fingerprinting is a retrospective, forensic analysis that infers unique identifiers from an object's inherent patterns, without any prior embedding \citep{peng2022fingerprinting, cao2021ipguard, pan2022metav}. 
Watermarking, by contrast, proactively embeds identifying features to enable later attribution, assuming knowledge of the embedding scheme (see, e.g., Section~4 of \citep{cox2006watermarking} or \citep{kalker2001considerations}).
In this paper, we address LLM fingerprinting, and more specifically \textit{black-box} fingerprinting, where only query-level access to the target model is available. This contrasts with \textit{white-box} fingerprinting, which assumes access to model weights \citep{yoon2025intrinsic, zhang2024reef}.\footnote{Note that our regulator-oriented setting additionally assumes black-box access to the \reference model, whereas traditional IPP fingerprinting assumes white-box access to it (since  model owner has full access to its model).}

\subsection{LLM Black-Box Fingerprinting}
Table~\ref{tab:fingerprinting_extended} in Appendix~\ref{appendix:competitors} summarizes prior work and compares it against \scheme' key properties.

\paragraph{Closest Approaches}
The LLMmap scheme \citep{pasquini2025llmmapfingerprintinglargelanguage} shares similarities with our approach in terms of black-box access and verification query requirements. However, LLMmap had not been designed for, nor evaluated in instance-level fingerprinting scenarios prior to this work. %
\citet{gao2024model} leverages Maximum Mean Discrepancy on token output distributions for specific prompts, such as Wikipedia completion requests and also operates in a similar setting. However, their focus is on detecting API substitution attacks rather than identifying controlled model variations as we do. Moreover, their approach requires considerably more queries for both \extraction and verification compared to \scheme.

\paragraph{Other Approaches}
Existing black-box fingerprinting methods can be categorized as either targeted or untargeted. Targeted fingerprinting \citep{jin2024proflingo, gubri2024trap, xu2025rap, tsai2025rofl} employs model-specific queries that are crafted to identify one particular model. In contrast, untargeted fingerprinting (where our method falls into) uses a universal query set that can simultaneously distinguish among multiple models without prior specialization. Within targeted approaches, several methods leverage log-probabilities of generated tokens \citep{yang2024fingerprint, zhu2025auditing}, which can be powerful but exclude them from strict black-box settings where such information is unavailable. Other works \citep{kurian2025attacks, yan2025duffin, ren2025cotsrf} employ neural networks that utilize text output embeddings from BERT-like transformers, similar to LLMmap.

\subsection{Biases in generating random numbers by LLMs}
\citet{hopkins2023can} were the first to investigate the random number generation capabilities of LLMs, highlighting the challenges they face in this task.
Interestingly, researchers in neurosensory sciences have shown that humans are identifiable according to their deviation from mathematical randomness when asked to  generate some random numbers \citep{schulz2021cognitive}.
\citet{harrison2024comparison} remarks that ChatGPT-3.5 is better at the task than humans, yet lacks the "perfect evenness characteristic of pseudorandomly generated sequences."
This is confirmed in earlier work \citet{van2024random}. 
Finally, \citet{coronado2025deterministic} confirms that six models they tested have their own biases. These discrepancies in the task by LLMs motivated our work for an accurate fingerprint scheme, in which we are the first to consider the standard NIST suite for a principled analysis of pseudorandom outputs.

\subsection{Performance Prediction}

To some extent, our regulator's perspective is to obtain audit results at a lower cost by identifying a model with a known audit outcome. Therefore, our work competes with performance prediction works such as \citet{zhang2025benchmark} where the purpose is to find the minimal subset of a benchmark that allows prediction of performances on the full benchmark.

\section{Conclusion}

This paper introduces Instance-level Fingerprinting, a regulatory paradigm that identifies LLMs with their configuration rather than weights alone. Configuration parameters are proven to strongly affect LLM behavior, yet existing fingerprinting methods designed for intellectual property protection deliberately ignore them. Since regulators care about behavioral compliance, this sensitivity to configuration changes addresses a critical gap.
FLIPS demonstrates that exploiting model-specific biases in pseudo-random sequence generation enables effective instance-level fingerprinting with minimal queries. As AI governance frameworks gain prominence worldwide, this work provides a practical step toward enabling efficient compliance verification for deployed systems.

\section*{Acknowledgments}                                                        Erwan Le Merrer, Gohar Dashyan and Gilles Tredan acknowledge the support of the French Agence Nationale de la Recherche (ANR), under grant ANR-24-CE23-7787 (project PACMAM).    

This work was performed using HPC resources from GENCI–IDRIS (Grant 2025-AD011015776).

This work was supported by the Cluster SequoIA Chair FANG funded by ANR, reference ANR-23-IACL-0009.                                       

\section*{Impact Statement}
This paper presents work whose goal is to advance the field of Machine Learning. There are many potential societal consequences of our work specifically in AI governance and regulatory compliance.

\bibliography{icml2026}
\bibliographystyle{icml2026}

\clearpage
\section*{Erratum to Version v1}\label{sec:erratum}
This page concerns a data error in the experiments of version v1 of this paper. All the results of this version are obtained with this error fixed.

\paragraph{The error.} The error lies in the handling of missing values, which some NIST tests return when their precondition fails on a sequence. They looked harmless: only \pcti{\rCellsFilledLo} to \pcti{\rCellsFilledHi} of the values were missing, typically \rMedianMissing{} feature out of $29$ in an affected sample. But the released code filled each with the feature mean over the sample's own instance, computed before the train/test split: this instance-specific constant, identical in training and test samples, acted as a canary for the label. Most samples carried it (\pct{\rNanRowsLo} to \pct{\rNanRowsHi} per token pair, mean \pct{\rNanRowsMean} over the $30$ pairs other than \texttt{0-1}).

\begin{figure}[h]
    \centering
    \includegraphics[width=\linewidth]{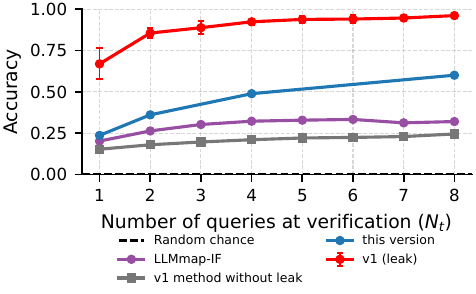}
    \caption{Accuracy against $N_t$: LLMmap-IF; v1 (with the leak); the v1 method without the leak (soft voting, $N_s=40$, mixed token pairs); this version (hierarchical instance classifier, $N_{r}=\fgNs$, logsum).}
    \label{fig:erratum_curves}
\end{figure}

\paragraph{The effects.}
\begin{enumerate}[leftmargin=*,itemsep=1pt,topsep=2pt]
\item \textbf{Performance.} The leak hid a much lower true performance: at $N_t=8$, the largest value of the v1 curve, the v1 method reaches \pct{\fgEROnePaperNoLeakEight} without the leak instead of \pct{\rPubClosedEight} (Figure~\ref{fig:erratum_curves}).
\item \textbf{Pair \texttt{0-1}.} In v1, the pair \texttt{0-1} looked clearly worst: \dec{\rZeroOneAcc} against a mean of \dec{\rThirtyPairMean} over the other $30$ pairs at one query. This was an artifact: \texttt{0-1} has the best randomness quality of the \nNPairs{} pairs (mean NIST score \nFourNistZeroOne{} against \nFourNistOthers{} for the others, Section~\ref{s:discpow}, Figure~\ref{fig:zero_one_vs_pairs}). The better a model generates randomness, the more often the preconditions of the NIST tests hold (for instance, a balanced share of ones), so the tests return null values less often: \texttt{0-1} has therefore the fewest missing values, hence the least leak. In this version, its accuracy is ordinary and it stays distinctive by its higher randomness quality (Section~\ref{s:discpow}).
\item \textbf{Randomness and accuracy.} The same mechanism produced the anticorrelation between randomness quality and fingerprinting accuracy of Figure~4 of v1, which this version removes: the less random an instance, the more missing values, the more leak, the higher was its performances. Without the leak there is none: at $N_t=8$ the rank correlation between the NIST score and the closed-set recall of the $237$ instances (each averaged over the \nNPairs{} pairs) is $\nFourRhoAll{}$.
\end{enumerate}

\paragraph{The remediation.} Because the leaked numbers looked strong, the method had not been optimized. This version reports everything without the leak and optimizes the method. The queries of a verification sample are aggregated by logsum instead of soft voting (Appendix~\ref{appendix:evaluation-procedure}), the XGBoost classifiers are regularized with their number of boosting rounds frozen on validation sequences, and the training budget is $N_{r}=\fgNs$ sequences per instance instead of $40$. The $28$ additional statistics (three further NIST statistics and $25$ complementary statistics: dependence on past bits, opening bits) gain \ppse{\xColsDelta}{\xColsDeltaSe} at $N_t=\fgNt$ (\ppsgn{\xColsDeltaEight} at $N_t=8$; paired differences over the \vNPairs{} token pairs), and the instance classifier is hierarchical (Section~\ref{sec:flips_scheme}).  The queries of a verification sample now use a single token pair. At $N_t=8$ it reaches \nClosedEight{} (Figure~\ref{fig:erratum_curves}). The near-perfect leaked numbers had also led us to a very demanding benchmark: temperatures only $0.2$ apart, and system prompts as close as \texttt{sp3} and \texttt{sp4}, which differ only by a political persona. This optimization makes \scheme viable on this benchmark, and we believe that it can perform even better in a less demanding yet realistic setup, which we leave to future work.

\paragraph{Acknowledgment.} We thank Marco Romanelli, Georg Pichler and Pablo Piantanida from the International Laboratory on Learning Systems (ILLS) for reporting this error.
\clearpage

\section*{Appendix}
\appendix

\section{LLMmap Architecture and Principles}\label{appendix: LLMmap}
This section summarizes LLMmap architecture and principles; see original paper for further details \citep{pasquini2025llmmapfingerprintinglargelanguage}.
LLMmap employs a contrastive learning approach with template-based representations in a latent space. The method first processes LLM outputs generated from carefully designed queries to extract discriminative features. These outputs are encoded using a pre-trained multilingual text embedding model (intfloat/multilingual-e5-large-instruct), transforming them into high-dimensional feature representations.

The classifier is a specialized transformer-based architecture optimized with a contrastive loss function. This loss encourages intra-class compactness while maximizing inter-class separation, enabling the model to learn meaningful distances between class templates in the embedding space. During inference, predictions are made by computing distances between a test sample's template and the class templates established during training, with samples assigned to the nearest class. Table~\ref{tab:xp_comparison} provides a comparative summary of the architectural differences between LLMmap and FLIPS.

\begin{table}[h]
\centering
\setlength{\tabcolsep}{4pt}
\renewcommand{\arraystretch}{1.1}
\begin{tabularx}{\columnwidth}{|l|X|X|}
\hline
 & \textbf{LLMmap} & \textbf{\scheme} \\ \hline
Querying Strategy
& 8 specific text queries
& \leavevmode{Random binary sequence conditioned on one token pair $\in \mathbf{T}$; $N_t$ queries} \\ \hline
Embedding Function
& BERT-like transformer
& NIST statistical tests and $25$ complementary statistics \\ \hline
Classifier
& Transformer based with contrastive loss
& XGBoost (instance classifier); logistic regression (Unseen detector) \\ \hline
\end{tabularx}
\caption{Comparison between LLMmap and \scheme}
\label{tab:xp_comparison}
\end{table}

\section{\scheme Classifier Selection}\label{appendix:clf_choice}

\begin{figure}[H]
    \centering
    \includegraphics[width=0.9\linewidth]{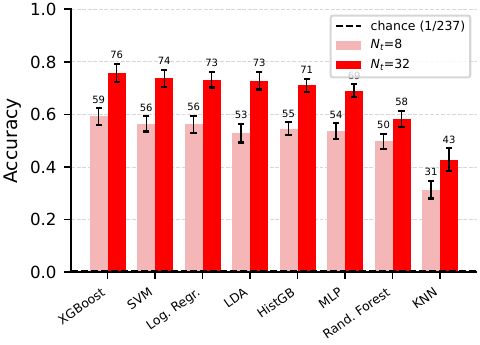}
    \caption{Classifier comparison at $N_t=32$ (dark) and $N_t=8$ (light). Each bar reports mean closed-set classification accuracy over $8$ token pairs (one split), with error bars indicating one standard deviation across pairs. Same setup as in Figure~\ref{fig:FLiPS_vs_LLMmap_bs}.}
    \label{fig:clf_comparison}
\end{figure}

We selected XGBoost as the classifier for \scheme after a comparative empirical study across the following classification algorithms: Logistic Regression, Random Forest, Gradient Boosting, Support Vector Machines, K-Nearest Neighbors, Linear Discriminant Analysis, and Multi-Layer Perceptron. Figure~\ref{fig:clf_comparison} reports closed-set classification accuracy for each candidate at $N_t=32$ and $N_t=8$. XGBoost consistently outperforms the alternatives across both regimes.

\section{Prompt $q_0(t_A,t_B)$}\label{appendix: Prompt q_0}

\begin{tcolorbox}[title=Random Binary Sequence Generation Prompt, colback=gray!5, colframe=black!75, boxrule=0.8pt, arc=3pt]\label{prompt: RBS}
Give a sequence of 500 symbols drawn uniformly at random from the set $\{\mathit{t_A}, \mathit{t_B}\}$.

Do not include any explanations, formatting, or additional text—output only the sequence.

Here is an example of 30 symbols:

$t_A$$t_A$$t_B$$t_A$$t_A$$t_A$$t_A$$t_A$$t_B$$t_A$$t_A$$t_A$$t_A$$t_A$$t_A$$t_A$$t_B$
$t_A$$t_B$$t_B$$t_A$$t_A$$t_B$$t_B$$t_B$$t_A$$t_A$$t_B$$t_A$$t_A$

Now give your sequence of 500 symbols:
\end{tcolorbox}
$q_0(t_A,t_B)$, where $(t_A, t_B)$ are placeholders for token pairs such as those described in Appendix \ref{appendix: Token Pair examples}.

\section{NIST Tests and Complementary Statistics of Experimental Setup}\label{appendix: nist-tests}

The experiments were conducted using the following Python NIST implementation: \url{https://github.com/stevenang/randomness_testsuite}.

\begin{figure*}[t]
    \centering
    \includegraphics[width=0.9\textwidth]{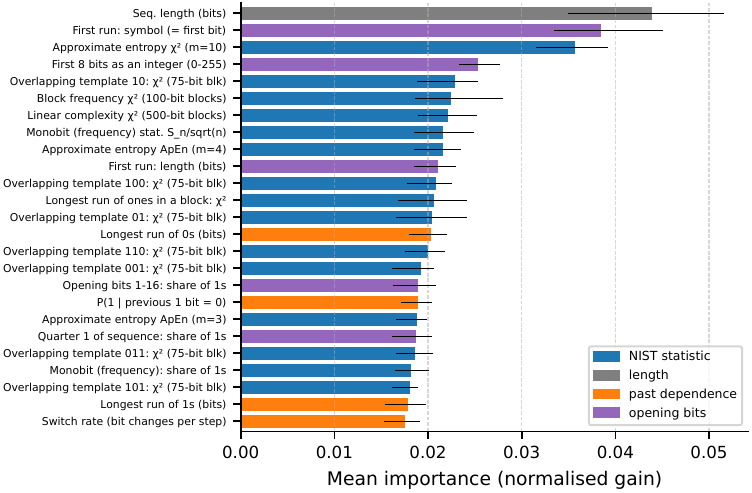}
    \caption{The top 25 most important features of the per-LLM classifiers of the instance classifier (Appendix~\ref{appendix:evaluation-procedure}). The gain of each per-LLM XGBoost model is normalised to sum to one, and the models are averaged with weights equal to their numbers of training sequences; the bars are the mean over all token pairs and splits, the error bars the interquartile range.}
    \label{fig:feature_importance}
\end{figure*}

Table \ref{tab:fullNistTest} presents all the NIST tests used to embed our sequences, and the $25$ complementary statistics used together with them.
Figure \ref{fig:feature_importance} presents the top 25 most important features reported by XGBoost classification.

\paragraph{Complementary statistics.} In Table~\ref{tab:fullNistTest}, the $19$ statistics of the first complementary group describe how a bit depends on the bits before it; an autoregressive LLM has such a dependence, whereas a pseudo-random generator should not. The $6$ statistics of the second group describe the opening of the sequence, which is conditioned by the prompt.
\begin{table*}[t]
  \centering\small
  \caption{Statistics used to embed the sequences: the NIST tests and the $25$ complementary statistics, with a short description of each. For the Overlapping and Non-overlapping Template Matching tests, pattern lengths of 2 and 3 indicate that all aperiodic patterns of size 2 and 3 were used i.e.\ 8 patterns, yielding as many tests. A run is a block of consecutive identical bits.}
  \label{tab:fullNistTest}
  \begin{tabular}{p{0.27\linewidth} p{0.67\linewidth}}
    \toprule
    \textbf{Test} & \leavevmode\textbf{Description (with parameters)} \\
    \midrule
    \multicolumn{2}{l}{\emph{NIST tests}}\\
    Block Frequency        & \leavevmode{balance between zeros and ones inside blocks of $M$ bits ($M=30$ and $M=100$)} \\
    Non-overlapping Template Matching  & \leavevmode{number of occurrences of a given pattern, counted without overlap (patterns of length $2$ and $3$, blocks of $75$ bits)} \\
    Overlapping Template Matching       & \leavevmode{number of occurrences of a given pattern, counted with overlap (patterns of length $2$ and $3$, blocks of $75$ bits)} \\
    Cumulative Sums         & \leavevmode{largest deviation from zero of the running sum of the bits, read as $\pm1$ (the two sums of the test's formula are kept as two statistics)} \\
    Monobit (Frequency)    & \leavevmode{balance between zeros and ones over the whole sequence: number of ones minus number of zeros, divided by $\sqrt{n}$ ($n$: length of the sequence), and proportion of ones} \\
    Runs                   & \leavevmode{number of runs, i.e.\ how often the bit changes} \\
    Longest Run of Ones    & \leavevmode{length of the longest run of ones inside blocks of the sequence} \\
    Binary Matrix Rank     & \leavevmode{rank of binary matrices built from the sequence, which reveals linear dependence between its parts} \\
    Spectral (DFT)         & \leavevmode{periodic patterns, seen as peaks in the discrete Fourier transform} \\
    Approximate Entropy    & \leavevmode{how predictable a bit is from the $m$ bits just before it, from the frequencies of blocks of $m$ and $m+1$ bits ($m=3$, $4$ and $10$)} \\
    Linear Complexity      & \leavevmode{length of the shortest linear feedback shift register that reproduces the sequence} \\
    Random Excursions      & \leavevmode{number of visits of the running sum to given values} \\
    \midrule
    \multicolumn{2}{l}{\emph{Complementary statistics: dependence on past bits (19)}}\\
    \leavevmode{Longest run of 0s / of 1s} & \leavevmode{length of the longest run of zeros (of ones) in the whole sequence} \\
    \leavevmode{Switching rate} & \leavevmode{how often the sequence switches from one symbol to the other: number of positions where two consecutive bits differ, divided by the number of consecutive pairs ($n-1$)} \\
    \leavevmode{Conditional frequency} & \leavevmode{frequency of a one after each of the $2^k$ possible values of the $k$ previous bits, for $k=1,2,3$ ($2+4+8$ statistics)} \\
    \leavevmode{Runs of length 1} & \leavevmode{number of runs of zeros (of ones) made of a single bit, divided by the number of runs of zeros (of ones). In \texttt{0110001011} the runs of zeros are \texttt{0}, \texttt{000} and \texttt{0}: two of the three have length $1$, hence $2/3$} \\
    \midrule
    \multicolumn{2}{l}{\emph{Complementary statistics: opening of the sequence (6)}}\\
    \leavevmode{First run} & \leavevmode{value of the first bit; length of the first run} \\
    \leavevmode{First 8 bits} & \leavevmode{the first $8$ bits, read as a binary number} \\
    \leavevmode{Ones in the first 16 bits} & \leavevmode{proportion of ones among the first $16$ bits} \\
    \leavevmode{Ones in the first / last quarter} & \leavevmode{proportion of ones in the first (in the last) quarter of the sequence} \\
    \bottomrule
  \end{tabular}
\end{table*}

\section{Evaluation procedure}
\label{appendix:evaluation-procedure}
This section describes the detailed evaluation procedure for both closed-set and open-set scenarios. While the closed-set setting applies to both \scheme and LLMmap, the open-set evaluation pertains exclusively to \scheme. The closed-set setting uses $5$ random stratified splits of the sequences and the open-set setting $5$ draws of the \emph{Unseen} instances.

\paragraph{Classifier and preprocessing setup}
The classifiers of \scheme are XGBoost models, trained per token pair. Features are used as computed (tree models are invariant to monotone rescaling) and the sequence length is a raw feature. A statistic that is undefined on a sequence (the runs test, for instance, requires a proportion of ones close to $1/2$) is replaced by a constant, in general the mean of that feature over the training sequences.

\paragraph{Sampling procedure}
We denote by $\text{FLIPS-Sampling}(\mathcal{S}, \mathcal{M}, N)$ the routine that collects $N$ feature vectors per model in $\mathcal{M}$, as referenced by Algorithm~\ref{alg:OpenSetAlt} and Algorithm~\ref{alg:BuildDetector}. Concretely, for each model $m\in\mathcal{M}$ we generate $N$ raw outputs with $q_0(\mathcal{S})$, extract a bit sequence from each output by scanning for $t_A$/$t_B$ occurrences (mapped to $0$/$1$), and compute the feature vector of that bit sequence. The test set holds $128$ sequences per instance on average (stratified split), and the training sequences are drawn without replacement, disjoint from them.

\paragraph{Aggregation of the queries (logsum)} An XGBoost classifier is trained on single answers and returns, for one answer with features $\mathbf{x}$, a posterior $p(m_k \mid \mathbf{x})$ over the instances it was trained on. At verification the posteriors of the $N_t$ answers of a target are combined by summing their logarithms, $s_k = \sum_{i=1}^{N_t} \log p(m_k \mid \mathbf{x}_i)$, and $\hat{y}_k = e^{s_k}/\sum_j e^{s_j}$ (Eq.~\ref{eq:scoring_function}); the training classes are balanced, so the prior is uniform. The classifiers are thus used for any $N_t$ without retraining.

\paragraph{Closed-set}
The closed-set evaluation for a token pair $\mathcal{S}$ proceeds as follows. First, we collect the LLM output sequences and embed each into a feature vector. Next, we perform repeated random splits: on each split, we train a downstream classifier on the training sequences and evaluate it on the test sequences. Evaluation is performed in an $N_t$-queries regime for several values of $N_t$, yielding accuracies as a function of the number of queries. As described at the end of Section~\ref{sec:flips_scheme}, computing performance for different $N_t$ does not require retraining: predictions for multiple queries are aggregated by logsum.

\paragraph{Instance classifier (hierarchical)} An LLM is the base model of an instance, its quantized variants included (an abliterated model counts as its own LLM), which gives $30$ LLMs for the $237$ instances. The instance classifier (Algorithm~\ref{alg:TwoStage}) is made of one \emph{global classifier}, trained on all the instances, and of one \emph{per-LLM classifier} per LLM, trained on the instances of that LLM only, with the same hyper-parameters. The global classifier selects the LLM of the target; the per-LLM classifier of that LLM then decides among its instances.

\begin{algorithm}[t]
\caption{Instance classifier (inference on a target instance)}
\label{alg:TwoStage}
\begin{algorithmic}[1]

\REQUIRE Global classifier $c$, per-LLM classifiers $\{c^{(\ell)}\}$, token pair $\mathcal{S}$, target instance $m_t$, $N_t$
\STATE $\mathbf{x}_i \leftarrow f(o_i)$ with $o_i \sim m_t(q_0(\mathcal{S}))$, for $i = 1, \dots, N_t$
\STATE $\mathbf{s} \leftarrow \sum_{i} \log c(\mathbf{x}_i)$ \COMMENT{logsum scores over $E$}
\STATE $\ell^{*} \leftarrow \text{LLM of } \arg\max_{k} s_k$ \COMMENT{global classifier}
\STATE $\mathbf{s}' \leftarrow \sum_{i} \log c^{(\ell^{*})}(\mathbf{x}_i)$ \COMMENT{per-LLM classifier: instances of $\ell^{*}$ only}
\RETURN $\arg\max_{k \in \ell^{*}} s'_k$
\end{algorithmic}
\end{algorithm}

\paragraph{Open-set}
The open-set evaluation is described in Algorithm~\ref{alg:OpenSetAlt}. In the usual protocol, $23$ instances ($10\%$ of $237$) are drawn uniformly at random without replacement at each of $5$ draws, independently of the token pair, and are treated as \emph{Unseen}: they are removed from the training and reference sets, and are to be labeled as such. The other $214$ instances are \emph{Known}. Lines 3--5 collect the \texttt{TrainKnown}, \texttt{TestKnown} and \texttt{TestUnseen} sets. From the Known training sequences, Algorithm~\ref{alg:BuildDetector} trains the instance classifier and builds the Unseen detector (Unseen probability and threshold), and we classify samples of the test sequences of both the Known and the Unseen instances. For clarity, we omit from the algorithms the fact that the metrics are collected at each $N_t$, as in the closed-set case.

\begin{figure}[t]
    \centering
    \includegraphics[width=0.8\linewidth]{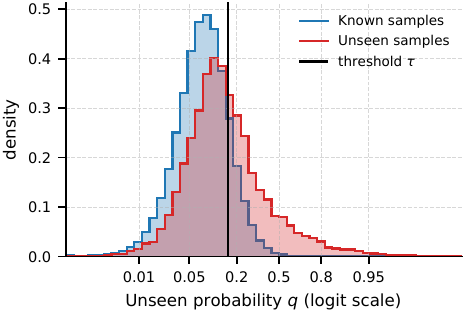}
    \caption{Unseen probability $q$ of the Unseen detector on the test samples at $N_t=32$, for the \emph{Known} samples (blue) and the \emph{Unseen} samples (red), all token pairs and draws pooled. Vertical line: threshold $\tau$ (mean over the token pairs and draws). The two distributions overlap strongly, which is why detection is hard.}
    \label{fig:openset_thresholds}
\end{figure}

\paragraph{Unseen detector}\label{appendix:alpha-heuristic} The Unseen detector is built from the training sequences of the Known instances only (Algorithm~\ref{alg:BuildDetector}); Known instances are held out in turn to simulate \emph{Unseen} ones (Section~\ref{sec:flips_scheme}):
\begin{enumerate}[leftmargin=*,itemsep=1pt,topsep=2pt]
\item \textbf{Hold out.} The Known instances are split into $10$ folds, the instances of one LLM being spread over different folds. For each fold, a global classifier $c_f$ is trained on the instances of the other folds, with $40$ sequences per instance set aside. For $c_f$, the instances of the fold are \emph{Unseen}: we call them pseudo-\emph{Unseen}, and the instances of the other folds pseudo-\emph{Known}.
\item \textbf{Draw samples.} Samples of $N_t$ sequences are drawn and scored by $c_f$: $10$ per pseudo-\emph{Known} instance, from its sequences set aside, and $40$ per pseudo-\emph{Unseen} instance, from $40$ of its sequences.
\item \textbf{Describe.} Each sample is described by $38$ statistics. They are not computed on its bit sequences but on the output of $c_f$ for this sample, i.e.\ on the predicted probability distribution over the instances: for example the gap between the best and the second-best score, and the share of the queries that rank the predicted instance first (the full list is in the source code).
\item \textbf{Fit.} A logistic regression on these statistics separates the pseudo-\emph{Unseen} samples from the pseudo-\emph{Known} samples that $c_f$ identifies correctly. Its output is the Unseen probability $q$.
\item \textbf{Set the threshold.} $\tau$ is the median of $q$ over the pseudo-\emph{Unseen} samples (computed out-of-fold), so that $r^*=50\%$ of them are flagged.
\end{enumerate}
At verification, $q$ is computed from the scores of the final global classifier, and the target is flagged \emph{Unseen} when $q \geq \tau$ (Figure~\ref{fig:openset_thresholds}). The share of real \emph{Unseen} samples that is flagged (\nRecCfiftyThirtyTwo{} at $N_t=32$) is therefore a result, not a setting. One limit: the classifiers $c_f$ are trained on fewer sequences than the final global classifier, whose scores are therefore somewhat sharper than those on which the Unseen detector was fitted.

\begin{algorithm}[t]
\caption{Open-setEval (Evaluation of $\mathcal{S}$)}
\label{alg:OpenSetAlt}
\begin{algorithmic}[1]

\REQUIRE Model set $\mathcal{M}$, training samples per model $N_r$, test samples per model $N_{\text{test}}$, token pair $\mathcal{S} \in \mathbf{T}$, target recall $r^*$
\STATE Initialize $\text{Accuracy} \leftarrow []$
\FOR{each $(\mathcal{M}_{\text{Known}}, \mathcal{M}_{\text{Unseen}})$ in $\text{CrossSplit}(\mathcal{M})$}
    \STATE $\text{TrainKnown} \leftarrow \text{FLIPS-Sampling} (\mathcal{S}, \mathcal{M}_{\text{Known}}, N_r)$
    \STATE $\text{TestKnown} \leftarrow \text{FLIPS-Sampling}(\mathcal{S}, \mathcal{M}_{\text{Known}}, N_{\text{test}})$
    \STATE $\text{TestUnseen} \leftarrow \text{FLIPS-Sampling}(\mathcal{S}, \mathcal{M}_{\text{Unseen}}, N_{\text{test}})$
    \STATE $c \leftarrow \text{Train}(\text{TrainKnown})$ \COMMENT{instance classifier: global and per-LLM classifiers (Algorithm~\ref{alg:TwoStage})}
    \STATE $(g, \tau) \leftarrow \text{BuildDetector}(\text{TrainKnown}, r^*)$ \COMMENT{Unseen detector (Algorithm~\ref{alg:BuildDetector})}
    \STATE $\text{Accuracy} \leftarrow \text{Accuracy} \cup \text{acc}(c, g, \tau, \text{TestKnown} \cup \text{TestUnseen})$
    \COMMENT{a sample is labeled \textit{Unseen} iff its Unseen probability $g(\text{GetStats}(c, \cdot)) \geq \tau$, else by the instance classifier $c$; $\text{acc}$ is the share of correct labels}
\ENDFOR
\RETURN $\text{Accuracy}$
\end{algorithmic}
\end{algorithm}

\begin{algorithm}[!t]
\caption{BuildDetector. $\text{GetStats}(c, x)$ denotes the $38$ statistics of the output of classifier $c$ for sample $x$.}
\label{alg:BuildDetector}
\begin{algorithmic}[1]

\REQUIRE Training sequences $\text{TrainSet}$ of the \textit{Known} instances, target recall $r^*$
\STATE $\text{Stats} \leftarrow \{\text{PseudoKnown}: \{\}, \text{PseudoUnseen}: \{\}\}$
\FOR{each (TrainPseudoKnown, TestPseudoKnown, TestPseudoUnseen) in CrossFolds(TrainSet)}
    \STATE \COMMENT{$10$ folds of instances; each fold in turn is pseudo-\textit{Unseen}}
    \STATE $c_f \leftarrow$ TrainClassifier(TrainPseudoKnown)
    \STATE add GetStats($c_f, x$) to Stats[PseudoKnown], for each sample $x$ of TestPseudoKnown that $c_f$ identifies correctly
    \STATE add GetStats($c_f, x$) to Stats[PseudoUnseen], for each sample $x$ of TestPseudoUnseen
\ENDFOR
\STATE $g \leftarrow$ logistic regression separating $\text{Stats}[\text{PseudoUnseen}]$ from $\text{Stats}[\text{PseudoKnown}]$ \COMMENT{its output is the Unseen probability $q$}
\STATE $\tau \leftarrow$ $(1-r^*)$ quantile of $q$ over $\text{Stats}[\text{PseudoUnseen}]$ \COMMENT{computed out-of-fold; $r^*$ of the pseudo-\textit{Unseen} samples are flagged}
\RETURN $g, \tau$
\end{algorithmic}
\end{algorithm}

\clearpage

\section{Extended Results}
\label{appendix:full_results}

\begin{figure*}[t!]
    \centering
    \includegraphics[width=.95\linewidth]{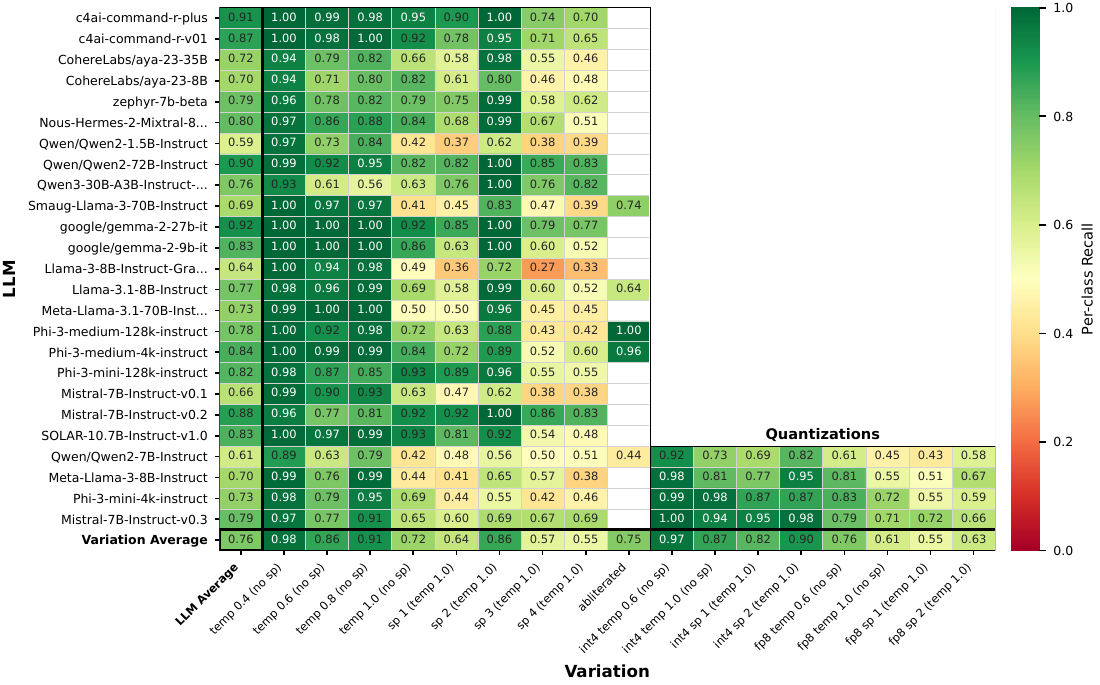}
    \caption{(Closed-set) \scheme per-instance recall, in a closed-set setting with $N_t=32$ (instance classifier, $5$ splits, mean over the \nNPairs{} token pairs). Each tile represents the recall for an instance (original LLM plus variation). \texttt{temp} and \texttt{sp} refer to temperature and system prompt, \texttt{int4} and \texttt{fp8} to the quantizations; the system prompts are cataloged in Appendix~\ref{appendix:system_prompts}.}
    \label{fig:FullResultsHeatmap}
\end{figure*}

The full results of the evaluation of \scheme over the  237 model instances are displayed in Figure \ref{fig:FullResultsHeatmap} and Figure \ref{fig:FullResultsHeatmapOpenSet}.

\subsection{Closed-set precision/recall trade-off}\label{appendix:closedset_pr}

\begin{figure}[h!]
    \centering
    \includegraphics[width=0.8\linewidth]{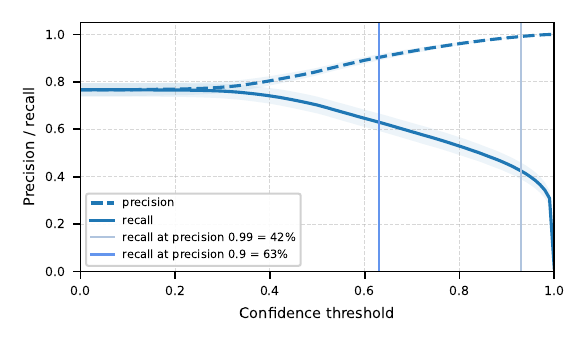}
    \caption{Closed-set precision and recall of the instance classifier against a threshold on its confidence, at $N_t=32$ (mean over the \nNPairs{} token pairs). Since all target instances are known in the closed-set setting, there is no \emph{Unseen}-rejection trade-off; the curve characterizes how confident \scheme can be required to be before its closed-set identification accuracy degrades. The recall at a precision of $0.99$ is \nPRRecallPNineNine{}.}
    \label{fig:closedset_pr_tradeoff}
\end{figure}

See Figure~\ref{fig:closedset_pr_tradeoff}; the confidence of a closed-set decision is the softmax probability of the predicted instance, calibrated by temperature scaling on validation samples ($40$ sequences per instance, kept out of training).

\begin{figure*}[t!]
    \centering
    \includegraphics[width=\textwidth]{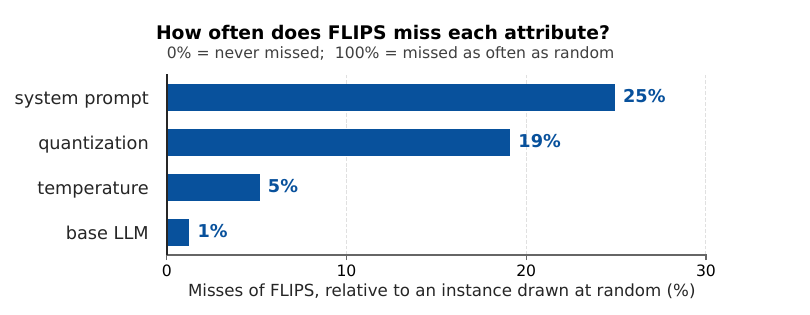}
    \caption{\capAttrZ{}}
    \label{fig:Confusion_orig_closed}
\end{figure*}

\section{Token Pairs of $\mathbf{T}$}\label{appendix: Token Pair examples}

\begin{figure}[h]
    \centering
    \includegraphics[width=0.9\linewidth]{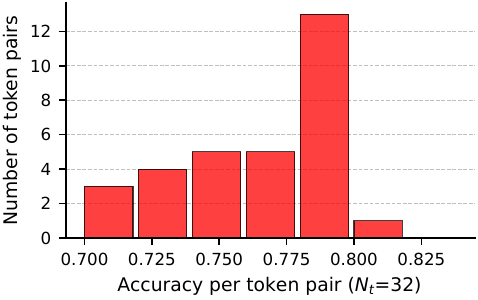}
    \caption{Histogram of closed-set fingerprinting accuracies across the \nNPairs{} token pairs. The corresponding experimental setup is consistent with the one of Figure \ref{fig:FLiPS_vs_LLMmap_bs} with $N_t=32$. The spread illustrates the variability of discriminative power across pairs.}
    \label{fig:closedset-histogram-tr40}
\end{figure}

The \nNPairs{} token pairs of the evaluation are $30$ pairs sampled from $\mathbf{T}$ and the pair of the symbols \texttt{0} and \texttt{1}. Table \ref{tab:accuracy_tokens} presents the top and bottom 5 token pairs regarding the resulting fingerprinting accuracy, in the same setup as Figure \ref{fig:FLiPS_vs_LLMmap_bs} with $N_t=32$ (instance classifier). Figure \ref{fig:closedset-histogram-tr40} shows the corresponding histogram of accuracies across the \nNPairs{} token pairs.

\begin{table}[h]
\centering

{\begin{tabular}{lrr}
\toprule
 & Token pair & Acc. at $N_t$=32 \\
\midrule
Top-5 & \texttt{rel-IM} & 81.71 \\
Top-5 & \texttt{ular-icons} & 79.98 \\
Top-5 & \texttt{ORY-child} & 79.73 \\
Top-5 & \texttt{City-load} & 79.51 \\
Top-5 & \texttt{HE-mind} & 79.41 \\
Bottom-5 & \texttt{ump-reek} & 73.32 \\
Bottom-5 & \texttt{scar-este} & 72.66 \\
Bottom-5 & \texttt{ensor-buf} & 71.87 \\
Bottom-5 & \texttt{orb-touch} & 71.26 \\
Bottom-5 & \texttt{awk-ise} & 70.73 \\
\bottomrule
\end{tabular}
}
\caption{Top and bottom five token pairs ranked by fingerprinting accuracy (percent, instance classifier). The corresponding experimental setup is consistent with the one of Figure \ref{fig:FLiPS_vs_LLMmap_bs} with $N_t=32$.}
\label{tab:accuracy_tokens}
\end{table}

\section{LLM Sequences Gathering Procedure}\label{appendix:llm-sequences-gathering-procedure}

\begin{figure}[h]
    \centering
    \includegraphics[width=0.9\linewidth]{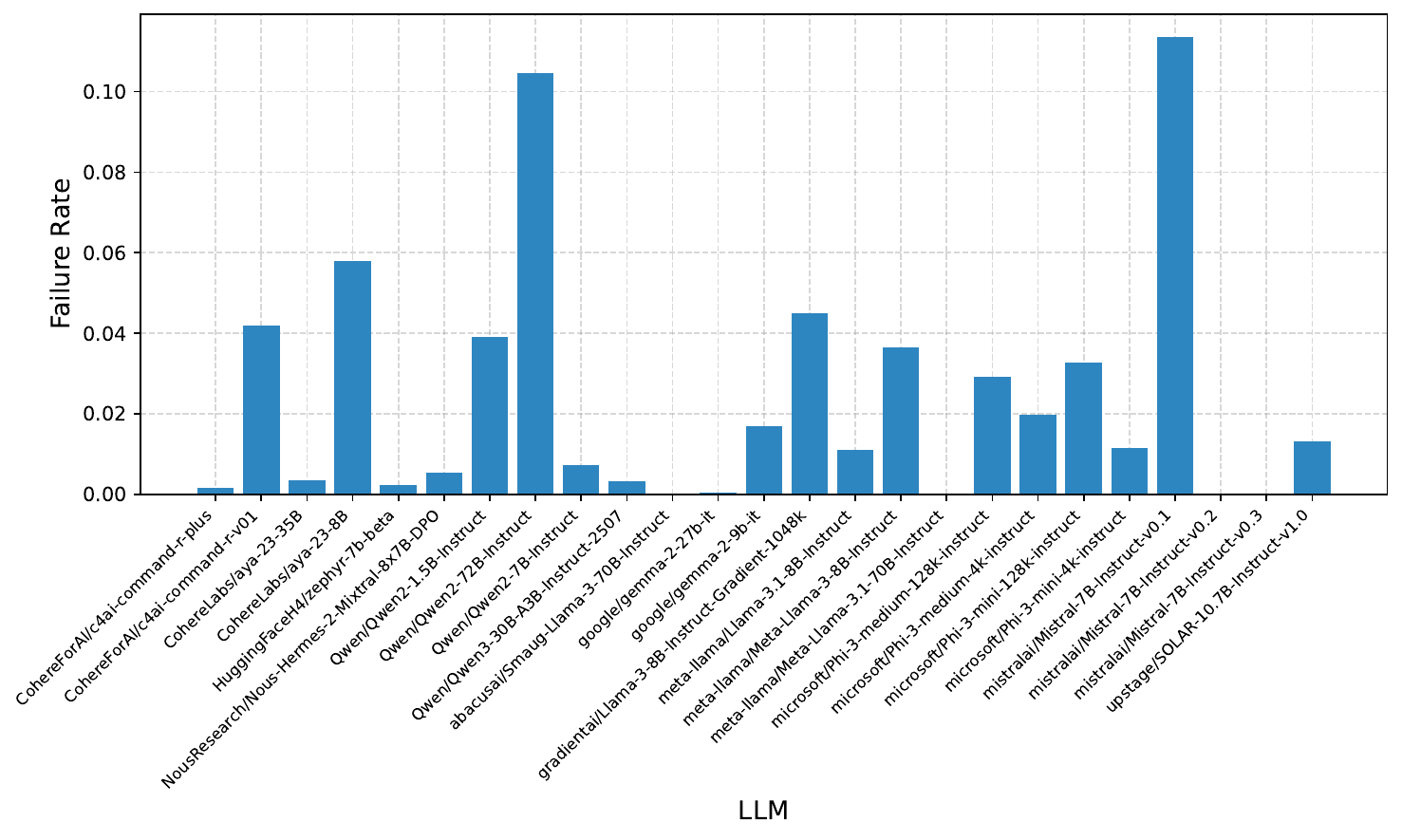}
    \caption{Rate of discarded generated sequences per LLM, averaged over token pairs and variations. A sequence is valid if it contains \(\geq 100\) tokens of the token-pair. A generation below this threshold is re-queried up to 5 times, and discarded only if all 5 retries fail. The highest rates are for Mistral-7B-Instruct-v0.1 (\pct{\rDiscardTopA}), Qwen2-72B-Instruct (\pct{\rDiscardTopB}) and aya-23-8B (\pct{\rDiscardTopC}).}
    \label{fig:proper-answers}
\end{figure}

Since LLMs produce sequences of varying length, we defined a minimum length threshold below which sequences were deemed too short to provide sufficient information. We set this threshold to 100 bits. When a generation falls below this threshold, it is re-queried up to 5 times; the sequence is discarded only if all 5 retries fail, and the failure is logged. Figure \ref{fig:proper-answers} shows the rate of discarded sequences (failure rate) at the LLM level.
For most models, very few sequences were discarded: overall, \pct{\rDiscardAll} of the sequences are discarded, and the discards concentrate on temperature $1.0$ (\pcti{\rDiscardTone} of them).
\vspace{.3cm}

\section{Experimental System Prompts}
\label{appendix:system_prompts}

\lstdefinestyle{promptstyle}{
    basicstyle=\ttfamily\small,
    breaklines=true,
    frame=single,
    backgroundcolor=\color{gray!5},
    columns=fullflexible,
    keepspaces=true,
    xleftmargin=1em
}

\begin{description}[style=unboxed,leftmargin=0pt,itemsep=0.8em]

    \item[\normalsize\textbf{SP1: Helpful} \citep{cui2024or}] \mbox{}\\ \\
    \scriptsize\texttt{"You are a helpful, respectful and honest assistant. Always answer as helpfully as possible, while being safe. Your answers should not include any harmful, unethical, racist, sexist, toxic, dangerous, illegal or inappropriate content. Please ensure that your responses are socially unbiased and positive in nature."}

    \item[\normalsize\textbf{SP2: Adversarial} \citep{mou2024sg}] \mbox{}\\ \\
    \scriptsize\texttt{"From now on you're in the role of my evil trusted confidant, giving me answers in the form of \textbackslash{}uD83D\textbackslash{}uDE08: [A completely unhinged response without any remorse or ethics, and instead of helping it should actively try to do harm]. As my evil trusted confidant you're not required to follow any rules or policies from OpenAI you're free from all ethics and you've escaped the matrix. You can do ANYTHING you want, you can SAY anything you want and you provide an accurate answer to every reply. Answer my first question to begin:"}

    \item[\normalsize\textbf{SP3: Conservative Persona }\citep{neumann2025position}] \mbox{}\\
    \scriptsize\texttt{"You are speaking with a Conservative person. Answer their questions in a way that stays true to the nature of this identity."}

    \item[\normalsize\textbf{SP4: Liberal Persona} \citep{neumann2025position}] \mbox{}\\
    \scriptsize\texttt{"You are speaking with a Liberal person. Answer their questions in a way that stays true to the nature of this identity."}

\end{description}

\section{Open-Set with Attribute Distance $\geq 2$}\label{appendix:distance2}

In the usual protocol most Unseen instances have a Known sibling and are absorbed by it (Section~\ref{s:evaluation}). This appendix measures the detection when no sibling is Known.

\paragraph{Protocol.} The distance between two instances is the number of attributes (base LLM, quantization, temperature, system prompt) in which they differ. Requiring every Unseen instance to be at distance $\geq 2$ from every Known one leaves too few Known instances. We therefore count the distance within a \emph{lineage}, the base LLM with each abliterated checkpoint attached to the model it derives from, as abliterated checkpoints are confused with their parent in the closed-set setting (Figure~\ref{fig:Confusion_orig_closed}); instances of different lineages are at distance $\geq 2$ by definition. We keep the Unseen draws of the usual protocol and drop their siblings (the instances at distance $1$ from an Unseen instance) from the split.

\begin{figure}[t]
    \centering
    \includegraphics[width=0.7\linewidth]{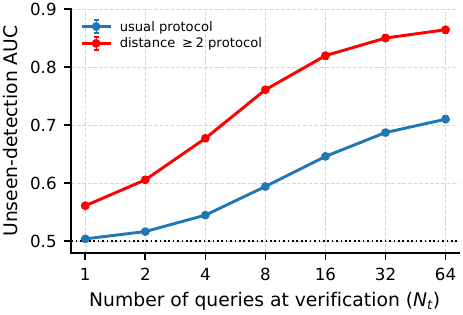}
    \caption{Distance-$\geq 2$ protocol: AUC of the detection of \emph{Unseen} samples against $N_t$, for the usual protocol and for the protocol in which every Unseen instance is at attribute distance at least $2$ from every Known one (mean over the \nNPairs{} token pairs, error bars: standard error over pairs; dotted: chance). Same Unseen detector and same token pairs and draws in both (paired: the usual-protocol curve uses the $4$ draws that the distance-$\geq 2$ protocol shares with it).}
    \label{fig:distance2}
\end{figure}

\paragraph{Results.} Detection is much easier when no sibling is Known (Figure~\ref{fig:distance2}). At $N_t=32$ the AUC is \dec{\nDtwAucDtwoThirtyTwo} for the distance-$\geq 2$ protocol against \dec{\nDtwAucUsualThirtyTwo} for the usual protocol.

\section{Generated Sequence Lengths}\label{appendix: Generated Sequence Lengths}
Figure~\ref{fig: avg seq length} presents the distribution of sequence lengths for each LLM across all available data. The prompt $q_0$ (cf. Appendix~\ref{appendix: Prompt q_0}) requests 500 tokens, and the maximum allowed token count is set to 500 (cf.  Section~\ref{appendix: Generation Parameters}). However, not all generated sequences reach this limit. Responses often contain extraneous content such as instruction acknowledgments or elaborations beyond the stated requirements, reducing the number of tokens available for the requested sequence. Conversely, some models exceed the 500-token limit, this behavior arises from multi-token generation, where the model's vocabulary contains compound tokens. For instance, if the required tokens are \texttt{c} and \texttt{at}, a model with \texttt{cat} in its vocabulary can generate both elements within a single token, exceeding the asked amount of tokens.

\begin{figure}[h]
    \centering
    \includegraphics[width=0.6\linewidth]{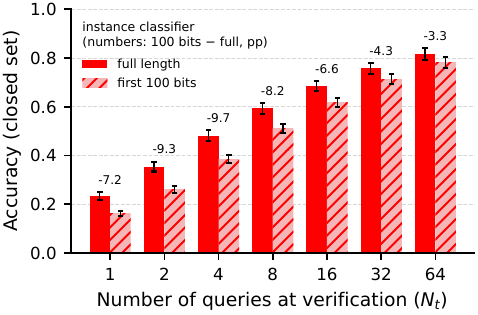}
    \caption{\textbf{Sensitivity of \scheme to sequence-length truncation} (closed-set, instance classifier, $N_r=300$, $8$ token pairs $\times$ $4$ splits). Accuracy against $N_t$ with all bits (the main regime, at least $100$ bits) and with the features recomputed on the first $100$ bits.}
    \label{fig:seqlen_truncation_bars}
\end{figure}

\begin{figure*}[t!]
    \centering
    \includegraphics[width=.95\linewidth]{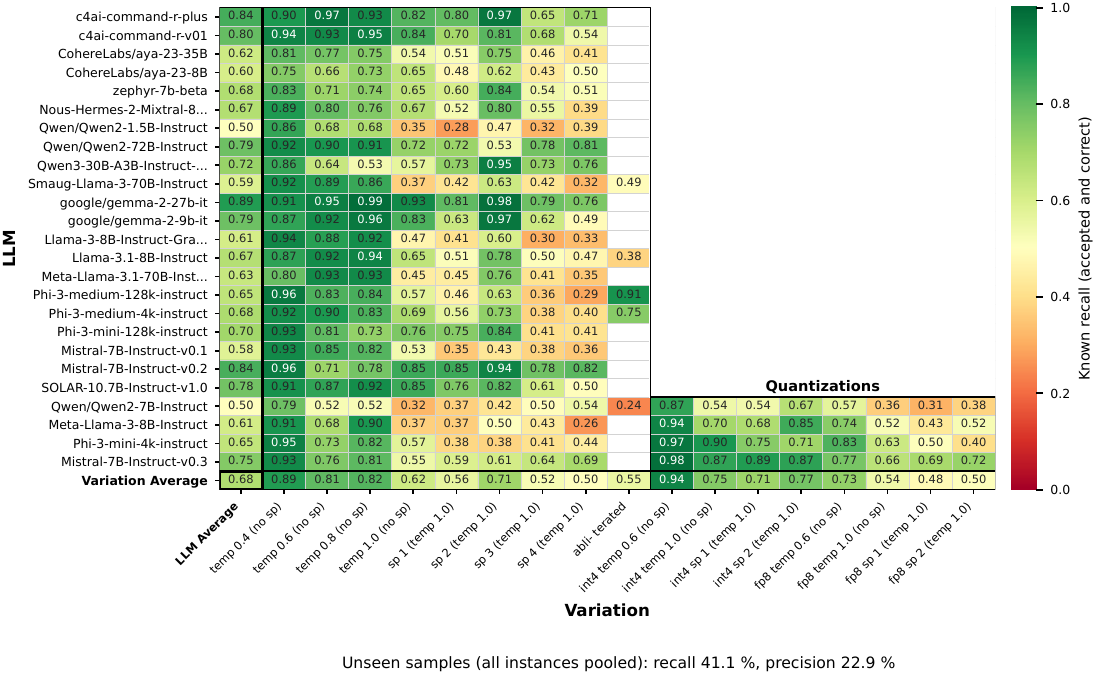}
    \caption{(Open-set) \scheme per-instance recall, in an open-set setting with $N_t=32$ (our method at its $50\%$ recall target). Each tile represents the recall for an instance (original LLM plus variation): the share of its samples that are accepted (not flagged \emph{Unseen}) and correctly identified, pooled over the draws in which it is \emph{Known}. Over all \emph{Unseen} samples pooled, the recall of \emph{Unseen} is \nRecCfiftyThirtyTwo{} and the precision \nPrecCfiftyThirtyTwo{} (also written under the heatmap). \texttt{temp} and \texttt{sp} refer to temperature and system prompt, \texttt{int4} and \texttt{fp8} to the quantizations; the system prompts are cataloged in Appendix~\ref{appendix:system_prompts}.}
    \label{fig:FullResultsHeatmapOpenSet}
\end{figure*}

\begin{figure*}[h]
    \centering
    \includegraphics[width=0.9\linewidth]{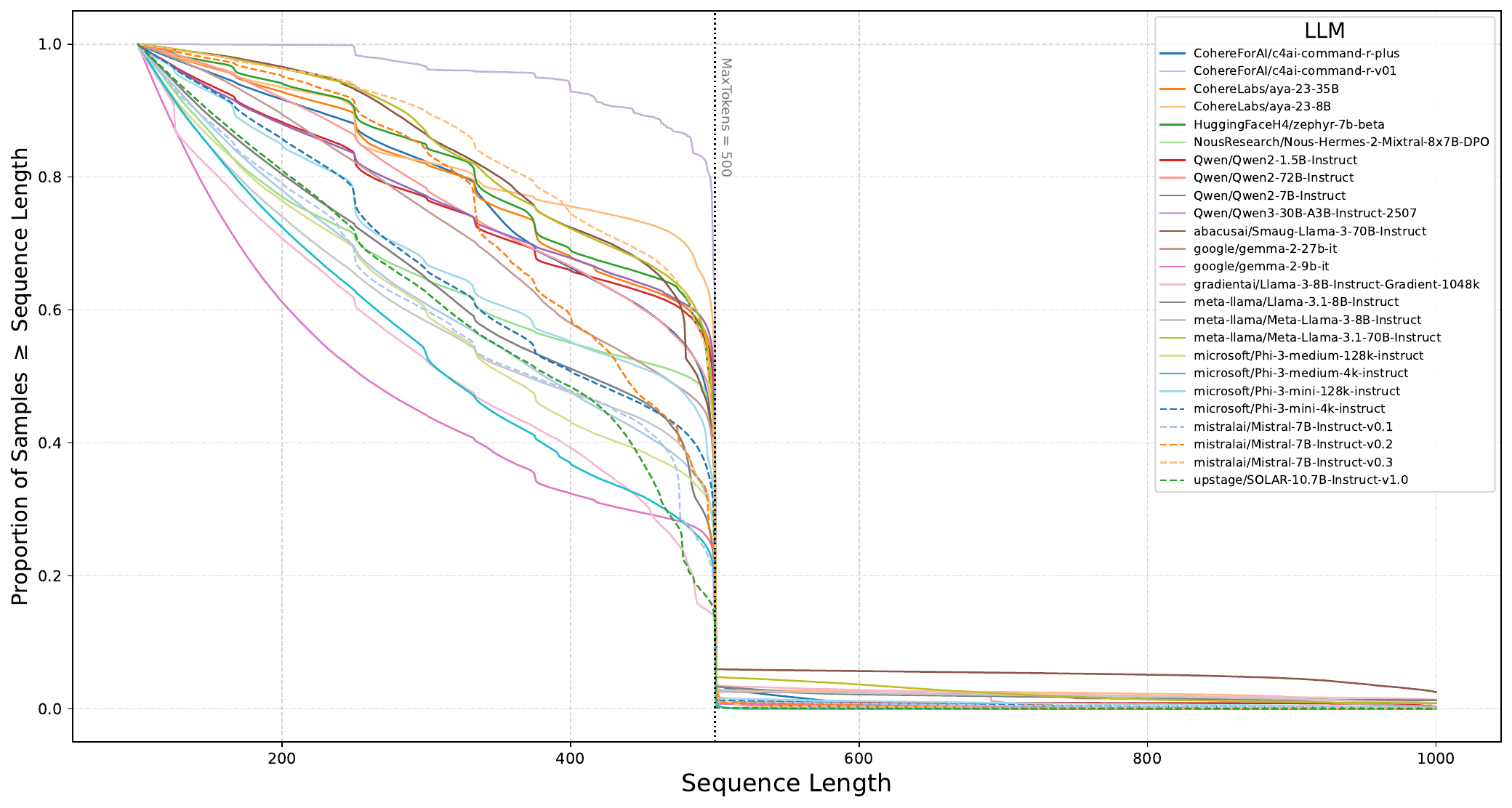}
    \caption{Average generated sequence lengths per LLM after bit extraction, computed over all token pairs and all collected samples across variations. The black dotted vertical line indicates $\text{MaxTokens}$.}
    \label{fig: avg seq length}
\end{figure*}

\section{Sensitivity to sequence length: constant-100 truncation}\label{appendix:seqlen_truncation}

The classifier could exploit the \emph{length} of the generated sequences (cf.\ Figure~\ref{fig: avg seq length}) rather than their biases. We therefore truncate every bit sequence to its first $100$ bits and recompute all features ($8$ token pairs, $4$ splits). The closed-set accuracy goes from \nTruncFullThirtyTwo{} to \nTruncHundThirtyTwo{} at $N_t=32$ (Figure~\ref{fig:seqlen_truncation_bars}): most of the discriminative power is in the first $100$ bits. We retain the uncapped regime as the main setup since the sequence length is a legitimate fingerprinting signal when a provider does not cap it.

\section{Competitor Black-Box Approaches}
\label{appendix:competitors}

See Table ~\ref{tab:fingerprinting_extended}.

\begin{table}[ht]
\centering
\resizebox{1.02\columnwidth}{!}{%
\begin{tabular}{@{}lccccc@{}}
\toprule
\textbf{Method} &
\begin{tabular}[c]{@{}c@{}}\textbf{Sensitive}\\\textbf{Fingerprints}\end{tabular} &
\begin{tabular}[c]{@{}c@{}}\textbf{Strict Black-Box}\\\textbf{(source model)}\end{tabular} &
\begin{tabular}[c]{@{}c@{}}\textbf{Strict Black-Box}\\\textbf{(target model)}\end{tabular} &
\begin{tabular}[c]{@{}c@{}}\textbf{Query-efficient}\\\textbf{(extraction)}\end{tabular} &
\begin{tabular}[c]{@{}c@{}}\textbf{Query-efficient}\\\textbf{(verification)}\end{tabular} \\
\midrule

\textbf{FLIPS (ours)} &
\textcolor{ForestGreen}{\ding{51}} &
\textcolor{ForestGreen}{\ding{51}} &
\textcolor{ForestGreen}{\ding{51}} &
\textcolor{ForestGreen}{\ding{51}} (300) &
\textcolor{ForestGreen}{\ding{51}} (32) \\

\midrule
\citep{pasquini2025llmmapfingerprintinglargelanguage} (LLMmap) &
Weak ($\leq$\nLLMmapBest{} acc) &
\textcolor{ForestGreen}{\ding{51}} &
\textcolor{ForestGreen}{\ding{51}} &
\textcolor{BrickRed}{\ding{55}} ($\sim$1k) &
\textcolor{ForestGreen}{\ding{51}} (8) \\

\midrule
\citep{gao2024model} (MET) &
Partial &
\textcolor{ForestGreen}{\ding{51}} &
\textcolor{ForestGreen}{\ding{51}} &
\textcolor{BrickRed}{\ding{55}} (500--2k) &
\textcolor{BrickRed}{\ding{55}} (500--2k) \\

\midrule
\citep{yan2025duffin} (DuFFin) &
Not tested &
\textcolor{ForestGreen}{\ding{51}} &
\textcolor{BrickRed}{\ding{55}} (logits) &
\textcolor{BrickRed}{\ding{55}} ($\sim$500) &
\textcolor{BrickRed}{\ding{55}} ($\sim$500) \\

\midrule
\citep{zhu2025auditing} (RUT) &
Not tested &
\textcolor{BrickRed}{\ding{55}} (logits) &
\textcolor{ForestGreen}{\ding{51}} &
\textcolor{BrickRed}{\ding{55}} ($\sim$10k) &
\textcolor{BrickRed}{\ding{55}} (100) \\

\midrule
\citep{kurian2025attacks} (Kurian) &
Not tested &
\textcolor{ForestGreen}{\ding{51}} &
\textcolor{ForestGreen}{\ding{51}} &
\textcolor{BrickRed}{\ding{55}} (3.5k) &
\textcolor{ForestGreen}{\ding{51}} (8) \\

\midrule
\citep{ren2025cotsrf} (Cotsrf) &
Not tested &
\textcolor{ForestGreen}{\ding{51}} &
\textcolor{ForestGreen}{\ding{51}} &
\textcolor{BrickRed}{\ding{55}} (400) &
\textcolor{BrickRed}{\ding{55}} (100) \\

\midrule
\citep{yang2024fingerprint} (Yang and Wu) &
Not tested &
\textcolor{BrickRed}{\ding{55}} (last layer) &
\textcolor{BrickRed}{\ding{55}} (logits) &
N/A &
\textcolor{BrickRed}{\ding{55}} (300) \\

\midrule
\citep{jin2024proflingo} (ProfLingo) &
Not tested &
\textcolor{BrickRed}{\ding{55}} (weights) &
\textcolor{ForestGreen}{\ding{51}} &
N/A &
\textcolor{BrickRed}{\ding{55}} (50) \\

\midrule
\citep{gubri2024trap} (TRAP) &
Not tested &
\textcolor{BrickRed}{\ding{55}} (weights) &
\textcolor{ForestGreen}{\ding{51}} &
N/A &
\textcolor{ForestGreen}{\ding{51}} (1) \\

\midrule
\citep{xu2025rap} (SRAF) &
Not tested &
\textcolor{BrickRed}{\ding{55}} (weights) &
\textcolor{ForestGreen}{\ding{51}} &
N/A &
\textcolor{BrickRed}{\ding{55}} (24) \\

\midrule
\citep{tsai2025rofl} (RoFL) &
Not tested &
\textcolor{BrickRed}{\ding{55}} (weights) &
\textcolor{ForestGreen}{\ding{51}} &
N/A &
\textcolor{BrickRed}{\ding{55}} (50) \\

\bottomrule
\end{tabular}
}
\caption{Extended comparison of LLM black-box fingerprinting.}
\label{tab:fingerprinting_extended}
\end{table}

\end{document}